\documentclass[anonymous]{llncs} %
\usepackage{graphicx}
\usepackage{algorithm}
\usepackage{algorithmic}
\usepackage{amsmath}
\usepackage{multirow}
\usepackage{booktabs}
\usepackage{subfigure}
\usepackage{hyperref}
\newcommand{\tabincell}[2]{\begin{tabular}{@{}#1@{}}#2\end{tabular}}
\usepackage{newfloat}
\usepackage{listings}
\usepackage{marvosym}
\usepackage{authblk}
\hypersetup{hypertex=true,
colorlinks=true,
linkcolor=blue,
anchorcolor=blue,
citecolor=blue}
\begin{document}

\title{ShapeWordNet: An Interpretable Shapelet Neural Network for Physiological Signal Classification}
\titlerunning{ShapeWordNet}
%\author{Submitted for Blind Review.} 
%\author{First Author\inst{1}\orcidID{0000-1111-2222-3333}
\author{Wenqiang He \inst{1} %\orcidID{0000-0002-4125-3421}
\and Mingyue Cheng \inst{1} %\textsuperscript{(\Letter)} 
\and Qi Liu \inst{1}\textsuperscript{(\Letter)} %\thanks{Corresponding Author.%This research was partially supported by grant from the National Natural Science Foundation of China (Grant No. 61922073). } %\textsuperscript{(\Letter)}
\and %\orcidID{2222--3333-4444-5555}} 
Zhi Li \inst{2} %\textsuperscript{(\Letter)}
}
\institute{Anhui Province Key Laboratory of Big Data Analysis and Application, University of Science and Technology of China, Hefei, China
\email{\{wenqianghe,mycheng\}@mail.ustc.edu.cn},
\email{qiliuql@ustc.edu.cn},
 \and Shenzhen International Graduate School, Tsinghua University, Shenzhen, China
 \email{zhilizl@sz.tsinghua.edu.cn}
}
%
%\authorrunning{Anonymous} %F. Author et al.
%\institute{Princeton University, Princeton NJ 08544, USA \and Springer Heidelberg, Tiergartenstr. 17, 69121 Heidelberg, Germany
%\url{http://www.springer.com/gp/computer-science/lncs} \and ABC Institute, Rupert-Karls-University Heidelberg, Heidelberg, Germany\\
\maketitle              % typeset the header of the contribution

\begin{abstract}
Physiological signals are high-dimensional time series of great practical values in medical and healthcare applications. However, previous works on its classification fail to obtain promising results due to the intractable data characteristics and the severe label sparsity issues. In this paper, we try to address these challenges by proposing a more effective and interpretable scheme tailored for the physiological signal classification task. Specifically, we exploit the time series shapelets to extract prominent local patterns and perform interpretable sequence discretization to distill the whole-series information. By doing so, the long and continuous raw signals are compressed into short and discrete token sequences, where both local patterns and global contexts are well preserved. Moreover, to alleviate the label sparsity issue, a multi-scale transformation strategy is adaptively designed to augment data and a cross-scale contrastive learning mechanism is accordingly devised to guide the model training. We name our method as ShapeWordNet and conduct extensive experiments on three real-world datasets to investigate its effectiveness. Comparative results show that our proposed scheme remarkably outperforms four categories of cutting-edge approaches. Visualization analysis further witnesses the good interpretability of the sequence discretization idea based on shapelets.
%Physiological signals are high-dimensional time series of great practical values in medical and healthcare applications. However, previous works on its classification fail to obtain promising results due to the intractable data characteristics and the severe label sparsity issues. In this paper, we try to address these problems and propose a more effective and interpretable scheme tailored for the physiological signal classification task. Specifically, we take advantage of the time series shapelets to extract prominent local patterns and conduct interpretable sequence discretization for information distillation. By doing so, the long and continuous raw signals are compressed into short and discrete token sequences, where both the local patterns and the global contexts are well preserved. In addition, to alleviate the label sparsity issue, a multi-scale data augmentation strategy is adaptively designed and equipped with a cross-scale contrastive learning mechanism for more effective representation learning. We name our method as ShapeWordNet and conduct extensive experiments on three real-world datasets to investigate its effectiveness. Comparative results demonstrate the significant outperformance of our proposed method over four categories of cutting-edge approaches. Visualization analysis further witnesses the good interpretability of the shapelet-based sequence discretization idea.

\end{abstract}
\section{Introduction}
%physiological data and APSC task
Physiological signal is an invaluable type of medical time series, which has broad applications in the healthcare domains, such as emotion recognition, seizure detection and heartbeat classification \cite{faust2018deep}. To effectively indicate the health state of human body, relevant information is often recorded simultaneously by multiple sensors through high-frequency and long-time sampling. For example, an electrocardiogram (ECG) signal recording can be sampled in 12 channels at a frequency of 500 HZ for at least 10 seconds to be used for diagnosing the cardiovascular condition of a patient. Nowadays, the fast progress of IoT spurs an explosive increase of physiological signals, making the traditional way of manually classifying such high-dimensional data not only costly but also inefficient \cite{krupinski2010long}. Hence, recent research has turned to artificial intelligence and machine learning for technical assistance \cite{che2018recurrent}.

Given the temporal data property, the physiological signal classification (PSC) task is often viewed as a typical time series classification (TSC) problem in the machine learning field, where a plentitude of TSC methods have been proposed and can be roughly grouped into two categories: the classical algorithms and the deep learning (DL) based approaches \cite{bagnall2017great,ismail2019deep}. 
Classical TSC methods focus on explainable feature engineering, where distinguishable features are designed and extracted from various perspectives. For instance, the ``golden standard'' 1-NN Dynamic Time Warping (DTW) paradigm \cite{ding2008querying,lines2015time} concentrates on comparing the similarity of global patterns, while the shapelet-based approaches \cite{ye2009time,grabocka2014learning,lines2012shapelet} aim at mining discriminative subsequences that maximally represent a class. Nevertheless, despite effective on small-scale and univariate datasets, classical methods do not scale well to the PSC task due to the difficulty of large-space feature selection and the inability to capture multi-variate interaction \cite{schafer2017multivariate}. 

In the past few years, deep learning based methods have reported incredible advancements in the TSC field, which avoid handcraft feature design and laborious feature selection by directly learning informative and low-dimensional representations from raw data \cite{zhang2020tapnet,ismail2019deep}. However, DL approaches require a large amount of labelled data to supervise model training, which is quite limited in the PSC scenario and may lead to performance degeneration. Besides, DL models provide little insight into the decisive factors and such black-box natures would impair their credibility in the healthcare field \cite{ahmad2018interpretable,lipton2018mythos}.
To overcome the above challenges and better adapt to the PSC task, one natural idea is to make full use of the strengths of both the classical and the deep learning based methods \cite{zhang2020tapnet}.

In this article, we propose a two-stage model named ShapeWordNet to provide a more effective  and interpretable solution to the PSC problem. To be specific, we novelly take advantage of the time series shapelets to extract discriminative local patterns as elementary ``words'' that contain certain class-relevant ``semantics'' of the original data. 
Then, based on these explainable words, we discretize the point-wise raw signal into a word-wise token sequence dubbed ShapeSentence for the whole-series information distillation, where we believe both significant local patterns and global contexts are well preserved. Moreover, in order to alleviate the label sparsity issue, the large shapelet candidate space is adaptively leveraged to augment the raw data in a multi-scale transformation way, and a cross-scale contrastive learning mechanism is accordingly constructed as an auxiliary objective to capture the data invariance. Finally, a scale-aware feature integrator is devised to fuse the representations of multi-scale ShapeSentences for class label prediction.

In summary, the main contributions of our work are as follows:
\begin{itemize}
    \item We propose an effective and interpretable scheme named ShapeWordNet tailored to the physiological signal classification task, which integrates the representation learning strengths of deep neural networks with the interpretability advantages of time series shapelets.
    \item We design a ShapeWord Discretization strategy to deal with the intractable data properties of physiological signals and devise a cross-scale contrastive learning mechanism to alleviate the label sparsity issues. To the best of our knowledge, this is the first work to utilize shapelets for explainable sequence discretization and deep learning's representational ability promotion.
    \item We conduct extensive experiments on three real-world datasets to investigate the effectiveness of ShapeWordNet. The comparative results validate the model's outperformance over four categories of TSC methods and the visualization analysis illustrates the good interpretability of the sequence discretization idea based on shapelets.
\end{itemize}

%\vspace{-0.5cm}
\section{Preliminaries}
\subsection{Problem Formulation}
Given a group of physiological signals  
$\mathcal{T} = \left\{ {{T_1},{T_2},...,{T_m}} \right\} \in {\mathcal{R}^{m \times d \times n}}$ and the corresponding label set 
$\mathcal{Y} = \left\{ {{y_1},{y_2},...,{y_m}} \right\} \in {\mathcal{R}^m}$, where each sample
${T_i} \in {\mathcal{R}^{d \times n}}$ is a $d$-dimensional sequence of $n$ time steps associated with a label $y_i$, the goal of physiological signal classification is to train a model ${f_\Theta }:T \mapsto y$ to predict the class label for a target instance.

\subsection{Definitions}
\subsubsection{Definition 1: Shapelet.}
%{\bfseries Definition 1: Shapelet.}
%{\itshape Shapelet.} 
A shapelet $\tilde S \in {\mathcal{R}^l}$ $(1 \le l \le n)$ is a type of subsequence that well discriminates classes \cite{ye2009time}. A good shapelet is supposed to have small $sDist$, i.e. the shapelet distance \cite{bagnall2015time}, to instances of one class and have large $sDist$ to those of another. The $sDist$ is defined as the minimum euclidean distance between $\tilde S$ and any subseries $w \in {W^l}$ of a given time series $T \in {\mathcal{R}^n}$ :
\begin{equation}
   \label{eq1}
    sDist(\tilde S,T) = {\min _{w \in {W^l}}}(dist(\tilde S,w)).
\end{equation}

\subsubsection{Definition 2: ShapeWord.}
%{\bfseries Definition 2: ShapeWord.}
%{\itshape ShapeWord.} 
A ShapeWord is defined as the cluster centroid of a set of similar shapelets, which represents the abstract prototype \cite{kolodner1992introduction} of their shared local pattern and can be referred to by the cluster label token $SW$:
\begin{equation}
	\label{eq2}
	%SW = Centroid(Cluster({{\tilde S}_1},...,{{\tilde S}_v})).
         \begin{aligned}
        ShapeWord &= ClusterCentroid({{\tilde S}_1},...,{{\tilde S}_v}), \\ 
         SW &= ClusterLabel({{\tilde S}_1},...,{{\tilde S}_v}).
         \end{aligned}
\end{equation}

\subsubsection{Definition 3: ShapeSentence.}
%{\bfseries Definition 3: ShapeSentence.}
%{\itshape ShapeSentence.} 
A ShapeSentence $SS$ is the discretized token sequence of the continuous raw series $T$, where each token $SW_i$ refers to a ShapeWord and $s$ is the length of this ShapeSentence:
\begin{equation}
	\label{eq3}
	T = \left[ {{t_1},...,{t_n}} \right] \in {\mathcal{R}^{d \times n}} \to SS = \left[ {S{W_1},...,S{W_s}} \right] \in {\mathcal{R}^{d \times s}}.
	%T = \left[ {{t_1},...,{t_n}} \right] \to SS = \left[ {S{W_1},...,S{W_s}} \right]
\end{equation}

\section{ShapeWordNet}
\begin{figure}[!htbp]
	\centering
        \vspace{-0.5cm}
	\includegraphics[width=\linewidth]{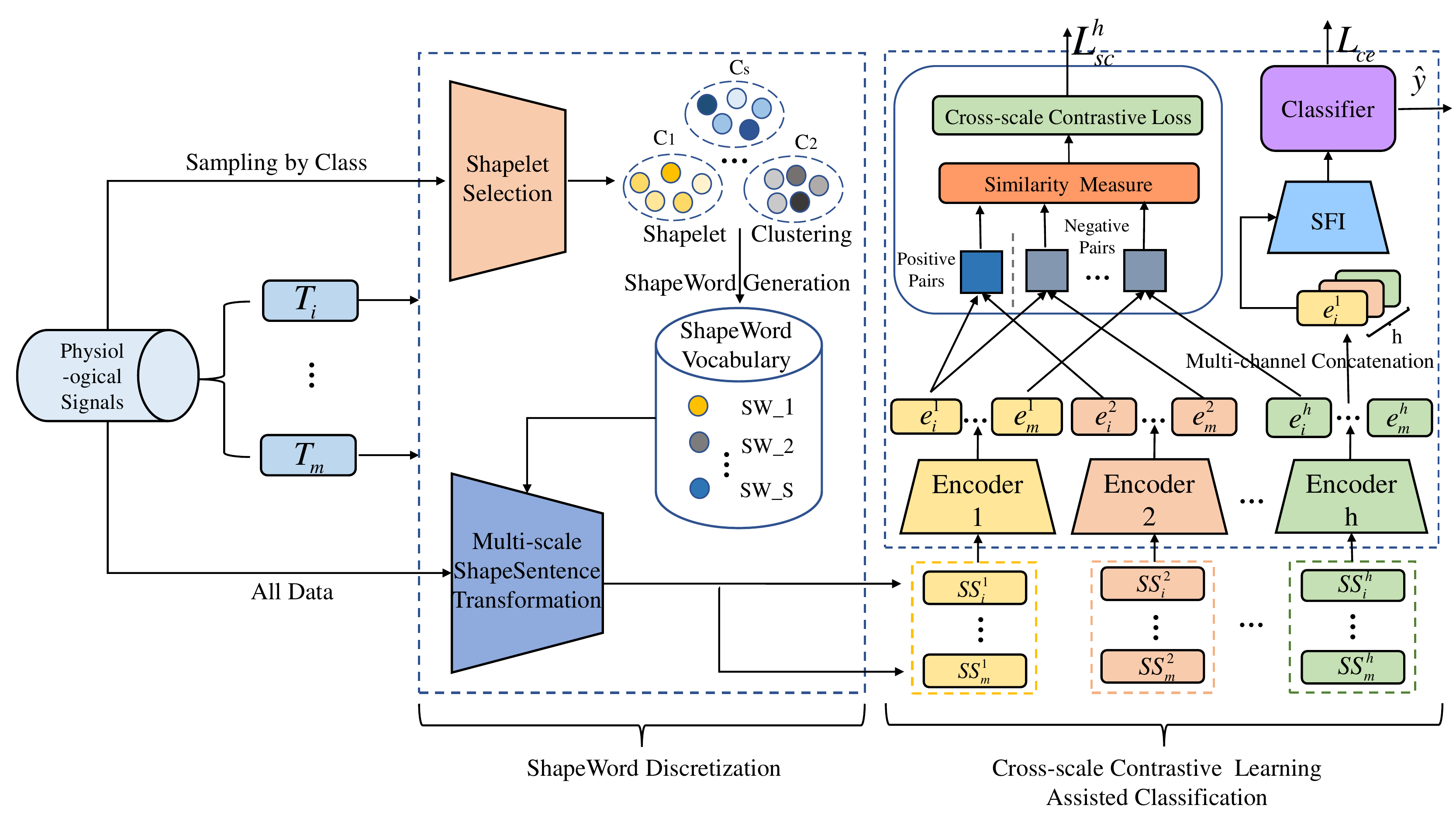}
	%\caption{An overview of the ShapeWordNet model, which consists of two stages:  (1) the ShapeWord Discretization including Shapelet Selection, ShapeWord Generation and Multi-scale ShapeSentence Transformation; (2) Cross-scale Contrastive Learning Assisted Classification involving Cross-scale Contrastive Learning, Scale-aware Feature Integration and Joint Optimization.}
        %\vspace{-0.8cm}
        \caption{An overview of the ShapeWordNet model.}
	\label{Fig_1}
\end{figure}

The overall architecture of our proposed ShapeWordNet is shown in Figure \ref{Fig_1}, which consists of two stages: the {\bfseries ShapeWord Discretization} stage and the {\bfseries Cross-scale Contrastive Learning Assisted Classification} stage.

\vspace{-0.8cm}
\subsection{ShapeWord Discretization}
The first stage includes three steps: (1) {\bfseries Shapelet Selection}, (2) {\bfseries ShapeWord Generation} and (3) {\bfseries Muti-scale ShapeSentence Transformation}.

\vspace{-0.4cm}
\subsubsection{Shapelet Selection.} Shapelets are discriminative subsequences that can offer explanatory insights into the problem domain \cite{ye2009time}. In this paper, we seize on such advantages of shapelets to extract interpretable and prominent local patterns. Traditional way of selecting shapelets is to evaluate the maximum information gain among all possible splits for each shapelet candidate, which would be extremely time-consuming in the physiological signal classification (PSC) scenario. To fast select shapelets, we combine the single-scan shapelet discovery algorithm \cite{lines2012shapelet} with a random sampling strategy.
Specifically, we first select 10 samples from each class at random. Then we generate shapelet candidates with a sliding window and evaluate each candidate's discrimination ability with the F-statistic measure that assesses the mean $sDist$ distribution differences between classes:
\begin{equation}
    \label{equation_4}
    F\left( S \right) = \frac{{{{\sum\nolimits_{v = 1}^V {{{\left( {{{\bar d}_{S,v}} - {{\bar d}_S}} \right)}^2}} } \mathord{\left/
 {\vphantom {{\sum\nolimits_{v = 1}^V {{{\left( {{{\bar d}_{S,v}} - {{\bar d}_S}} \right)}^2}} } {\left( {V - 1} \right)}}} \right.
 \kern-\nulldelimiterspace} {\left( {V - 1} \right)}}}}{{{{\sum\nolimits_{v = 1}^V {\sum\nolimits_{j = 1}^{{N_v}} {{{\left( {{{d}_{S,v,j}} - {{\bar d}_{S,v}}} \right)}^2}} } } \mathord{\left/
 {\vphantom {{\sum\nolimits_{v = 1}^V {\sum\nolimits_{j = 1}^{{N_v}} {{{\left( {{{d}_{S,v,j}} - {{\bar d}_{S,v}}} \right)}^2}} } } {\left( {N - V} \right)}}} \right.
 \kern-\nulldelimiterspace} {\left( {N - V} \right)}}}},
\end{equation}
where $V$ is the class number, $N$ is the total sample number, $N_v$ is the number of class $v$, $S$ is the shapelet candidate to be assessed, ${\bar d}_S$ is the mean value of its $sDist$ vector $D_S$, and ${d}_{S,v,j}$ is its $sDist$ with the $j$-th sample of class $v$.  

\subsubsection{ShapeWord Generation.} Although plenty of shapelets can be easily found, many of them are similar to each other, which brings about feature redundancy and increases computational complexity. To alleviate this issue, we propose to generate the prototypes \cite{kolodner1992introduction} that contain the key information shared by similar shapelets as the elementary units for sequence discretization. Toward this end, we cluster the selected shapelets with K-means and define the cluster centroids as their prototypes \cite{qiao2022interpretable}, as is suggested in Equation \ref{eq2}. Those prototypes are named ShapeWord and assigned with numeric cluster label tokens for reference. For the multivariate case, we simply repeat the aforementioned algorithm and generate ShapeWords for each variable. In doing so, we establish a \textit{vocabulary} of ShapeWords which encompasses the significant local patterns over all variable dimensions, as is illustrated in Figure \ref{Fig_1}. 

\begin{figure}[!htbp]
\centering 
\setlength{\belowdisplayskip}{3pt} 
\setlength{\abovedisplayskip}{3pt}
	\includegraphics[width=0.5\linewidth]{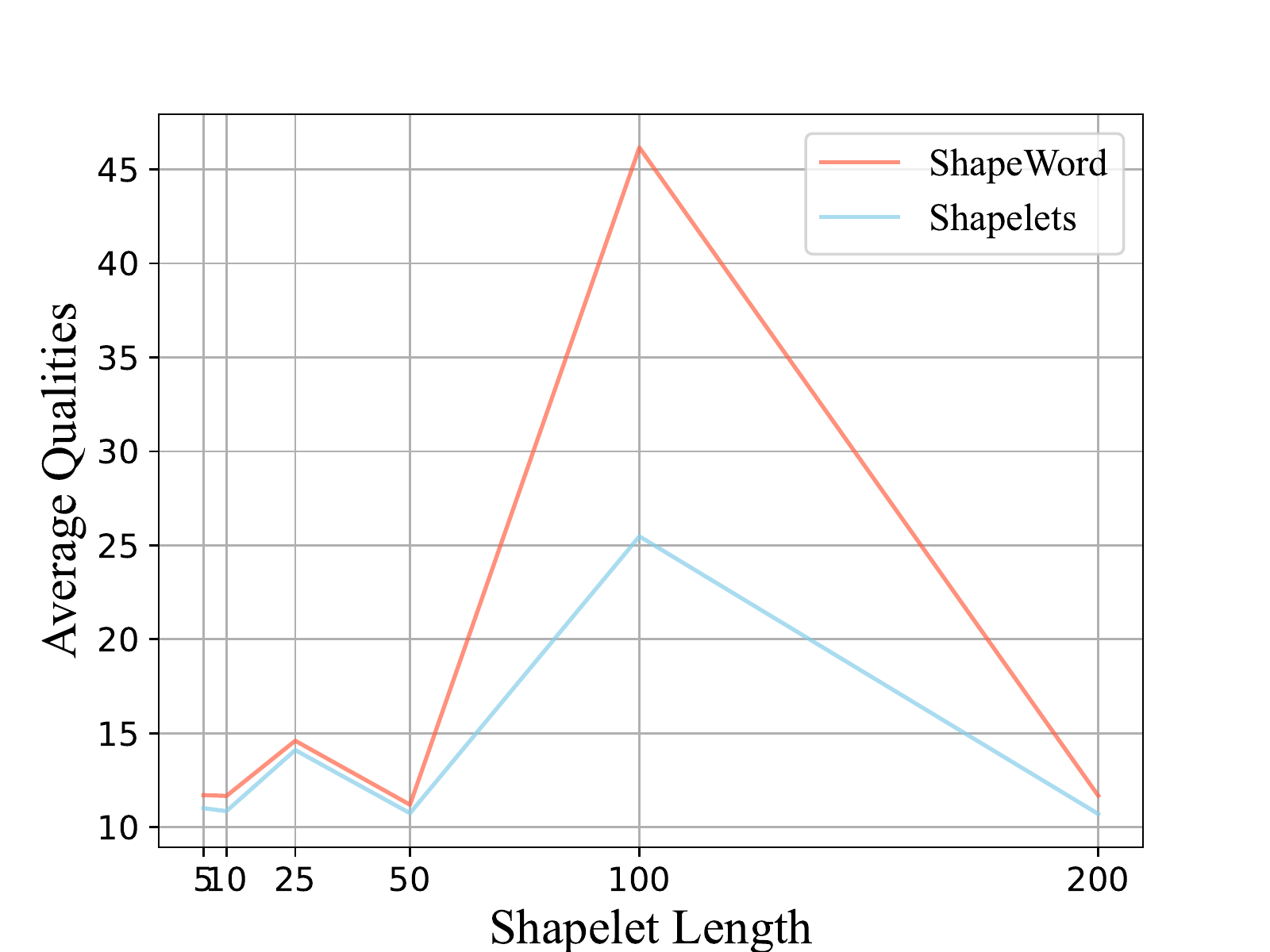}
	\caption{Results of the average F-statistic quality comparison between ShapeWords and the top-100 Shapelets w.r.t. the shapelet lengths, which is the higher the better. }
	\label{Fig_2}
\end{figure}

Moreover, to validate the discriminative edges of ShapeWords, we conduct a comparison experiment between the selected shapelets and the generated ShapeWords on the Sleep dataset \cite{867928}. In our experiment, we first generate shapelet candidates of different lengths from 5 to 200. For each scale, we select the top-100 shapelets to produce corresponding ShapeWords via K-means, where the cluster number is set equal to the class number, i.e. $N_{SW}=8$. Then, we establish a validation set containing 10 random samples of each class to compare the average F-statistic qualities of the ShapeWords with that of top-100 shapelets. The results in Figure \ref{Fig_2} show that the mean F-statistic scores of the ShapeWords are competitively higher regardless of scales, which concretely demonstrate the effectiveness of ShapeWords in representing the prototypes of similar shapelets.

\vspace{-0.7cm}   %调整图片与上文的垂直距离 
\vspace{-1cm}

\subsubsection{Multi-scale ShapeSentence Transformation.} 
Traditional shapelet-based methods focus on extracting local patterns for classification, while in the PSC scenario, the global contextual information such as the periodicity and variation of local patterns is also of critical importance. For example, the sinus arrhythmia can be more effectively diagnosed from a periodic perspective. Hence, based on the extracted local patterns represented by ShapeWords, we discretize the entire sequence to further distill the global contexts.
Firstly, we segment the original signals into non-overlapping subsequences via a sliding window of the ShapeWord size. Then, we assign each segment the cluster label token that refers to its nearest ShapeWord according to the Euclidean Distance. 
In doing so, a long and complex point-wise time series can be interpretably compressed into a much shorter and simpler word-wise token sequence, which preserve the key features both locally and globally and can be more robust to noise disturbance \cite{schafer2015boss}. 
We call such a token sequence ShapeSentence to suggest it is like a meaningful sentence in the natural language, where the whole sentence and its constituent words contain information at different semantic levels.

In addition, the ShapeSentence is quite scalable and can be adaptively extended as a data augmentation strategy to mitigate data sparsity issue. To be specific, we make the best of the large shapelet candidate space to generate different sizes of ShapeWords and transform each sample into multiple scales of ShapeSentences, which are further utilized to construct self-supervised signals in the contrastive learning mechanism. We dub this data augmentation technique as the \textbf{M}ulti-scale \textbf{S}hapeSentence \textbf{T}ransformation (MST) strategy and summarize its procedure in Algorithm \ref{alg2}. When $max=min+1$, it is the simplest single-scale ShapeSentence transformation version mentioned above.

\vspace{-0.2cm} 
\begin{algorithm}[htbp]
     \setlength{\belowdisplayskip}{10pt}
     \setlength{\abovedisplayskip}{10pt}
\begin{algorithmic}[1]
	\caption{\textbf{M}ulti-scale \textbf{S}hapeSentence \textbf{T}ransformation (MST) }
	\label{alg2}
	%\KwIn{Distribution over mete-training tasks: $p(\mathcal{T})$; Meta-testing tasks: $\mathcal{T}_{mt}$; Task-learning rate: $\alpha_{1}$; Meta-learning rate: $\alpha_{2}$.}
	%\KwOut{Labels of nodes in query set of $\mathcal{T}_{mt}$.}  
	%\BlankLine
        \STATE \textbf{Input:} sample set $T$, ShapeWord scale range $[min,max]$, ShapeWord vocabulary $SWList$ \\
        \STATE $AugmentedData \leftarrow \emptyset$  \\
	\FOR{ ${T_i}$ in $T$}
           \label{forins1}
           \STATE $MShapeSentences \leftarrow \emptyset$  \\
           \FOR{$l\leftarrow min$ to $max$}
              \label{forins2}
              
        \STATE ${W_{i}^{l}} \leftarrow segmentSequences\left( {{T_i},l} \right)$
             
             \FOR{subsequence $S$ in $W_{i}^{l}$}
              \label{forins3}
          \STATE ${SW_h} \leftarrow findClosest\left( {S,SWList} \right)$
           \STATE $MShapeSentences.append\left( {SW_h} \right)$
           \ENDFOR
    
    \STATE $AugmentedData.append\left( MShapeSentences \right)$
    \ENDFOR
    \ENDFOR
    \STATE \textbf{return} $AugmentedData$
\end{algorithmic}
\end{algorithm}
%\vspace{-1cm}
%merge the two graphs

\vspace{-0.7cm}
\subsection{Cross-scale Contrastive Learning Assisted Classification}
The second stage adopts the paradigm of multi-task learning, where the model training is assisted by the \textbf{Cross-scale} \textbf{Contrastive} \textbf{Learning} and the learnt representations are fused through \textbf{Scale-aware Feature Integration} before final classification.

\vspace{-0.5cm}
\subsubsection {Cross-scale Contrastive Learning.} Self-supervised learning has emerged as an alternative paradigm to overcome deep learning's heavy dependence on manual labels by leveraging the input data itself as supervision \cite{liu2021self}. In recent years, great breakthroughs in this field has been achieved by the contrastive learning \cite{chen2020simple}, which aims at ``learning to compare” through the Noise Contrastive Estimation (NCE) \cite{gutmann2010noise} or the InfoNCE objectives \cite{van2018representation}. In this work, we adpatively design a cross-scale contrastive learning mechanism to alleviate the label sparsity issues of PSC by constructing the self-supervised signals based on the multi-scale transformed ShapeSentences.  
Since a pair of large-scale and small-scale ShapeSentences of the same sample can be regarded as its observations from multi-scale perspectives, it is safe to hypothesize that there exists latent invariance behind them which we can enable the feature encoders to capture \cite{hou2021stock}. With this intuition, we encode each scale of ShapeSentence into fixed-size representations and compute the InfoNCE loss by comparing their similarities. 

As is illustrated in Figure \ref{Fig_1}, for instance, $h$ scales of sample $i$'s ShapeSentences, i.e. $SS_i^1,SS_i^2,...,SS_i^h$, are first fed into a set of encoders $E{n^1}\left(  \cdot  \right),...,E{n^h}\left(  \cdot  \right)$ to obtain their representations, i.e. $e_i^1 = E{n^1}\left( {SS_i^1} \right),...,e_i^h = E{n^h}\left( {SS_i^h} \right)$ (one encoder corresponds to one scale).
Then, in order to make the encoders capable of capturing the invariance shared by different scales of ShapeSentences, we define the cross-scale contrastive loss ${{L}_{sc}^h}$ as:
\begin{equation}
	\label{equ_5}
	L_{sc}^h = \frac{1}{{h \choose 2}}\sum\limits_{u = 1}^{h - 1} {\sum\limits_{v = u + 1}^h {L_{u,v}^{sc}}} ,
\end{equation}
\begin{equation}
	\label{equ_6}
	L_{u,v}^{sc} =  {E_{\left( {{e_i^u},{e_i^v}} \right) \sim {P_{u,v}^i}}}\left[ - {\log \frac{{f\left( {e_i^u,e_i^v} \right)}}{{f\left( {e_i^u,e_i^v} \right) + \sum\limits_{i \ne j} {f\left( {e_i^u,e_j^v} \right)} }}} \right] ,
\end{equation}
\begin{equation}
	f\left( {e_i^u,e_i^v} \right) = {{{{\left( {e_i^u} \right)}^T}e_i^v} \mathord{\left/
			{\vphantom {{{{\left( {e_i^u} \right)}^T}e_i^v} \tau }} \right.
			\kern-\nulldelimiterspace} \tau },
\end{equation}
where ${L_{u,v}^{sc}}$ is the contrastive loss between the representations of the $u$-scale ShapeSentences and the $v$-scale ShapeSentences, with ${{P_{u,v}^i}}$ as their joint sample distribution and $f\left(  \cdot  \right)$ being the representation similarity measure. In this article, we simply apply the vector inner product to compute representation similarities, view representation pair $(e_i^u,e_i^v)$ from the same sample as positive, and randomly select $N-1$ different samples from the marginal distributions of other samples within the same one mini-batch as negative, e.g. the negative ${e_j^v}$ from the $v$-scale marginal distribution of another sample $j$, as is suggested by InfoNCE \cite{van2018representation}. 

In our scheme, we choose the deep dilated causal convolutional neural network \cite{franceschi2019unsupervised} as the encoder backbone given its high efficiency and outstanding excellence in capturing long-range dependencies \cite{BaiTCN2018}. Besides, to learn the multivariate interactions, we input different variable's ShapeSentence into different channels of the encoder to obtain the representation feature vectors \cite{zheng2014time}. 

\vspace{-0.5cm}
\subsubsection{Scale-aware Feature Integration for Classification.} Before classification, we need to integrate the learned representations from different scales of ShapeSentences at first. In terms of multi-scale feature fusion, general methods like average pooling, maxpooling and direct concatenation \cite{cui2016multi} are all model-agnostic that do not take the domain particularities into consideration. To adaptively make full use of the multi-scale information in our method, we regard each scale's representation as complementary to the raw data's invariant features and concatenate them by channel. We then input the concatenated representation tensor into the the \textbf{S}cale-aware \textbf{F}eature \textbf{I}ntegrator (SFI) for feature fusion: 
\begin{equation}
        \label{integrator}
	{C_i} = SFI\left( {{E_i}} \right),
\end{equation}
where SFI is a single one-dimensional convolution layer with different channels catering to different scales of ShapeSentence representations, ${E_i} = \left[ {e_i^1,...,e_i^h} \right] \in {\mathcal{R}^{p \times h}}$ is sample $i$'s concatenated representation tensor, and ${C_i} \in {\mathcal{R}^q}$ is the integrated feature vector.
%and ${c_i} \in {\vmathbb{R}^q}$ is the output feature of SFI

Finally, we input the fused feature representation ${C_i}$ into a single linear classifier to obtain the classification outcome ${\hat y}_i$:
\begin{equation}
       \label{classify}
	%{{\hat y}_i} = {W_c}  {c_i},
       {{\hat y}_i} = {W_c}^\top{C_i} + {W_0},
\end{equation}
where $W_c \in {\mathcal{R}^{c \times q}} $ and ${W_0} \in {\mathcal{R}^c}$ are learnable parameters. 

\subsubsection{Multi-task Optimization.} To optimize the whole network, we combine the cross-entropy loss for classification with the cross-scale contrastive loss for self-supervision as a multi-task goal \cite{ruder2017overview} for joint training:
\begin{equation}
   \label{eq_10}
   \begin{aligned}
		L & = {L_{ce}} + \lambda L_{sc}^h \\ 
		& =  - \sum\limits_c^{\left| C \right|} {{y_c}\log \left( {{{\hat y}_c}} \right)}  + \frac{\lambda }{{h \choose 2}}\sum\limits_{u = 1}^{h - 1} {\sum\limits_{v = u + 1}^h {L_{u,v}^{sc}} }  \\ 
  \end{aligned} ,
\end{equation}
where $\lambda$ is used to balance different losses and $h$ is the number of scales.

\vspace{-0.3cm}
\section{Experiments}
\iffalse
In this section, we conduct extensive experiments to answer the following research questions:
\begin{itemize}
    \item {\bfseries RQ1:} Does our proposed ShapeWordNet outperform the state-of-the-art TSC methods on the APSC task?
    \item {\bfseries RQ2:} Do the ShapeWord Discretization stage, the Cross-scale Contrastive Learning Mechanism and Scale-aware Feature Integration component really work?
    \item {\bfseries RQ3:} How do we explain the way ShapeWord Discretization works?
    \item {\bfseries RQ4:} How do the hyper-parameters like the number of scales, the $\lambda$ in Eq \ref{eq_10}, the ShapeWord Number and the ShapeWord Length influence the performance?
\end{itemize}
\fi

\subsection{Experimental Setup} 
%\subsubsection{Evaluation Tasks.} 

\subsubsection{Datasets.} We conduct experiments on three real-world public datasets from PhysioNet \cite{goldberger2000physiobank}, where two datasets are ECG signals used for cardiac disease classification in the 2020 Physionet/Computing Cardiology
Challenge  \cite{alday2020classification} and one dataset contains EEG signals popular in sleep-stage classification \cite{867928}. 
In our experiment, we pick out part of the single-label samples and randomly split them into 80\%-20\% train-test datasets. The statistics of these datasets are summarized in Table \ref{table_1}, where Trainsize/Testsize means the sample number of the training/testing dataset, Dim/Len refers to the variable number and time steps, and Ratio stands for the ratio of one class number to all. %of a class's sample size to the largest one's.     

 %\vspace{-0.2cm}
\begin{table}[!htbp]
  \centering
  \caption{Descriptive statistics of three datasets.}
    \label{table_1}
  \setlength{\tabcolsep}{5mm}{
   \begin{tabular}{cccc}
   	\toprule
   	\multirow{2}*{Property} &\multicolumn{3}{c}{Datasets} \\
   	\cline{2-4}
   	&CPSC &Georgia &Sleep \\
   	\midrule
   	Triansize &5,123 &2,676 &12,787 \\
   	Testsize &1,268  &668 &1,421\\
   	%Variables  &12  &12 &2 \\
   	%Timesteps  &5,000 &5,000 &3,000 \\
   	Dim/Len &12/5,000 &12/5,000 &2/3,000 \\
   	%Classes &9 &9 &8 \\
   	Category &ECG &ECG &EEG \\ 
   	%Classes/Category &9/ECG &9/ECG &8/EEG \\
   	%Type/Labels &ECG/9 &ECG/9 &EEG/8 \\
   	\midrule
   	\tabincell{c}{Multi-class \\ Ratio (\%)} & \tabincell{c}{[14.07, 24.24, \\ 15.42, 9.53, \\ 12.26, 10.85, \\ 3.03, 2.69, 7.91]} &\tabincell{c}{[52.73, 12.97, \\ 7.47, 6.50, \\ 6.69, 3.81, \\ 3.77, 3.51, 2.54]} &\tabincell{c}{[48.42, 3.65, \\ 21.40, 4.25, \\ 4.40, 9.93, \\ 0.08, 7.86]} \\
   	%\tabincell{c}{Multi-class \\ RCR} & \tabincell{c}{[0.5805, 1.0000,\\ 0.6361, 0.3929,\\ 0.5056, 0.1111,\\ 0.1248, 0.4477,\\ 0.3261]} &\tabincell{c}{[1.0000, 0.2459, \\ 0.1417, 0.1233, \\ 0.1269, 0.0723,\\ 0.0716, 0.0666, \\ 0.0482]} &\tabincell{c}{[1.0000, 0.0784, \\ 0.4352, 0.0898,\\ 0.0897, 0.2084,\\ 0.0016, 0.1671]} \\
   	\midrule
   	\tabincell{c}{Binary \\ Ratio (\%)} & [14.07, 85.93] & [52.73, 47.27] & [48.42, 51.58] \\
   	%\tabincell{c}{Binary \\ RCR} & [0.1638, 1.0000] & [1.0000, 0.8965] & [0.9339, 1.0000] \\
   	\bottomrule
   \end{tabular}
   }
\end{table}

\vspace{-0.3cm}
\subsubsection{Baselines and Variants.} We compare three variants of our ShapeWordNet (SWN) scheme with four categories of time series classification (TSC) baselines:

{\bfseries (1) Shapelet-based:} We employ the Shapelet Transformation (\textbf{ST}) \cite{lines2012shapelet} and Learning Shapelets (\textbf{LS}) \cite{grabocka2014learning} for the first baseline group. ST searches the shapelets and transforms the original data into distance vectors, while LS learns the shapelets directly by optimizing the goal function of classification.

{\bfseries (2) Dictionary-based:} We pick out the classical \textbf{SAX-VSM} \cite{senin2013sax} and the recent \textbf{WEASEL+MUSE} \cite{schafer2017multivariate} for sequence discretization comparison, which respectively utilize the SAX words and SFA words for time series discretization and build classifiers based on their frequency patterns.
%\textbf{BOSSVS} \cite{schafer2016scalable} and

{\bfseries (3) SOTA:} This group contains two state-of-the-art TSC methods, including the the non-DL model MiniRocket \cite{dempster2021minirocket} and the DL model TapNet \cite{zhang2020tapnet}.

{\bfseries (4) CNN-based:} Considering the noticeable achievements of convolutional neural networks (CNN) in sequence modeling \cite{BaiTCN2018}, we adopt four different architectures of CNN-based TSC approaches as the deep learning baselines, which are \textbf{MCNN} \cite{cui2016multi}, \textbf{LSTM-FCN} \cite{karim2017lstm}, \textbf{ResCNN} \cite{zou2019integration} and \textbf{TCN} \cite{BaiTCN2018}.

{\bfseries (5) Variants:} We put forward three variants of our method for ablation study: (1) \textbf{SWN w/o SD} stands for the ShapeWordNet without ShapeWord Discretization, i.e. the encoder backbone and the \textbf{TCN} baseline, 
(2) \textbf{SWN w/o CCLM} represents the ShapeWordNet without the Cross-scale Contrastive Learning Mechanism, and (3) \textbf{SWN w/o SFI} means the ShapeWordNet without the Scale-aware Feature Integrator.

\vspace{-0.5cm}
\subsubsection{Implementation Details. } 
 We set the $layer\_depth=3$, $kernel\_size=3$, and $out\_channels=50$ as the default parameters for each of the dilated causal convolutional encoders. In terms of the ShapeWord/shapelet/word number parameters in three SWN variants, Dictionary-based and Shapelet-based baselines, we set them equal to the task class number times the variable dimensions of each dataset, i.e. 8*2 for Sleep and 9*12 for CPSC and Georgia. As for the ShapeWord/shapelet/word lengths, we set 10 for SWN w/o CCLM, SAX-VSM, WEASEL+MUSE, ST and LS, and set [10,25,50] for SWN w/o SFI and SWN. In SWN w/o SFI and SWN, the $\lambda$ in Equation \ref{eq_10} to balance loss is set to be 0.5. Besides, we set the $batch\_size=30$, $training\_epochs=50$ and use Adam optimizer with $learning\_rate=0.001$ for all methods.

We evaluate the classification performance with two metrics: \textbf{Accuracy (ACC)} and \textbf{Macro F1-score (MAF1)}. ACC can measure the overall performance by calculating how many samples are correctly classified in total, while MAF1 can avoid the measurement bias caused by class imbalance and assess a model's discrimination ability more fairly.

We have implemented the proposed method in python 3.7 and run all the experiments on a machine of CentOS 7.9.2009 with 4 Tesla V100 and 2 Intel Xeon Gold 5218 @2.30GHz. 

\vspace{-0.2cm}
\subsection{Performance Comparison}
To comprehensively evaluate the performance of our method, we conduct two experimental tasks of binary classification (BC) and multi-class classification (MC) on each dataset, where all the labels in MC other than the positive ones constitute the negative labels in BC. According to Table \ref{table_1}, the BC class ratios of Georgia and Sleep are more balanced than their MC class ratios, which enables BC tasks to serve as the control experiments concerning the label sparsity issue. Although the BC class ratio of CPSC is less balanced than MC class ratio, CPSC has a more balanced MC ratio than Georgia and Sleep, making it a control dataset to indicate the influence of class balance on model performance. 
Table \ref{table_2} reports the ACC and MAF1 of different methods on two tasks times three datasets and denotes the best ACC and MAF1 for each task with boldface. Consistent with our intuition, the major observations are summarized as follows:

%\vspace{-0.5cm}   %调整图片与上文的垂直距离  
\begin{table}[!htbp]
     \setlength{\belowdisplayskip}{2cm}
    \centering
    \caption{The performance comparison between different methods on BC and MC tasks of three datasets with ACC and MAF1 metrics. The best performing methods are boldfaced and the best beaselines are underlined. Improv (\%) measures the relative improvements of SWN variants over the best baselines. Note that the improvements are statistically significant with a two-sided t-test $p \textless 0.01$.  }
        \label{table_2}
	%\vspace{-0.2cm}
        %\setlength{\tabcolsep}{0.4mm}
        \resizebox{\linewidth}{!}{ %#此处！表示根据根据宽高比进行自适应缩放
		\begin{tabular}{ccccccccccccc}
			\toprule
			%\multirow{2}*{Category} & 
			\multirow{3}*{Method} &\multicolumn{4}{c}{CPSC} &\multicolumn{4}{c}{Georgia} &\multicolumn{4}{c}{Sleep} \\
			\cmidrule(r){2-5}  \cmidrule(r){6-9} \cmidrule(r){10-13} 
			%&\multicolumn{2}{c}{Binary Classification} &\multicolumn{2}{c}{Multi-class Classification}&\multicolumn{2}{c}{Binary Classification} &\multicolumn{2}{c}{Multi-class Classification}&\multicolumn{2}{c}{Binary Classification} &\multicolumn{2}{c}{Multi-class Classification}\\
			&\multicolumn{2}{c}{2 Classes} &\multicolumn{2}{c}{9 Classes} &\multicolumn{2}{c}{2 Classes} &\multicolumn{2}{c}{9 Classes} &\multicolumn{2}{c}{2 Classes} &\multicolumn{2}{c}{8 Classes} \\
			%\noalign{\smallskip}
			\cmidrule(r){2-3}  \cmidrule(r){4-5} \cmidrule(r){6-7} \cmidrule(r){8-9}
			\cmidrule(r){10-11} \cmidrule(r){12-13}
			
			&ACC &MAF1 &ACC &MAF1 &ACC &MAF1 &ACC &MAF1 &ACC &MAF1 &ACC &MAF1\\
			\midrule
			%\multirow{2}*{Shapelet\_based} & 
			LS & 0.8400 &0.4579 & 0.1554 &0.0299 &0.4970 &0.4970 &0.2814 &0.1112 &0.7847 &0.7843 &0.6946 &0.3244\\
			ST & 0.6500 &0.5911 & \underline{0.6900} &0.1331 &0.5100 &0.4516 &0.2500 &\underline{0.2319} &0.8300 &0.8311 &0.5600 &0.2839\\
			\midrule
               SAX-VSM &0.1617 &0.1426 &0.1483	&0.1141	&0.5973	&\underline{0.5904}	&0.1272	&0.1106	&0.6868	&0.6666	&0.5771	&0.2994 \\
               WEASEL+MUSE	&0.8303	&0.4987	&0.2505	&0.3850	&\underline{0.6102}	&0.5561	&\underline{0.5653}	&0.1567	&0.7136	&0.7126	&0.5517	&0.3334 \\
                \midrule
               MiniRocket &0.8450	&0.4646	&0.0935	&0.0220	&0.5308	&0.3802	&0.0793	&0.0164	&\underline{0.9369}	&\underline{0.9363}	&0.6864	&0.4211 \\
               TapNet	&0.7167	&0.5366	&0.1372	&0.1148	&0.5132	&0.5123	&0.2695	&0.1180	&0.7910	&0.7904	&0.5742	&0.2775 \\
                \midrule
			%BOSSVS &0.1562 &0.1355 &0.0489 &0.0456 &0.4865 &0.3298 &0.0404 &0.0145 &0.6791 &0.6327 &0.1956 &0.1258\\
			%SAX-KNN &0.7886 &0.5741 &0.1909 &0.1066 &0.5030 &0.4984 &0.4431 &0.0841 &0.5468 &0.5097 &0.4687 &0.1807\\
			MCNN &0.8564 &0.4613 &0.1044 &0.0210 &0.4850 &0.3266 &0.4850 &0.0726 &0.5011 &0.3338 &0.5039 &0.0838\\
			LSTM-FCN &\underline{0.8967} &\underline{0.7865} &0.6727 &\underline{0.6134} &0.5150 &0.3399 &0.1617 &0.0607 &0.7607 &0.7566 &0.6144 &0.2489\\
			ResCNN &0.8431 &0.4574 &0.1300 &0.0489 &0.5225 &0.4976 &0.1766 &0.0417 &0.7741 &0.7575 &0.1323 &0.0535 \\
			TCN (SWN w/o SD) &0.8904 &0.7734 &0.6491 &0.5423 &0.5928 &0.5883 &0.5045 &0.1740 &0.8951 &0.8946 &\underline{0.7241} &\underline{0.4645} \\
			\midrule
			SWN w/o CCLM &0.9014 &0.7893 &0.6924 &0.6395 &\textbf{0.7590} &\textbf{0.7590} &0.7246 &0.3650 &0.9198 &0.9192 &0.76425 &0.5210\\
			SWN w/o SFI &0.9085 &0.8059 &0.7397 &0.6778 &0.7350 &0.7325 &0.7156 &\textbf{0.4538} &0.9346 &0.9343
			&0.8023 &0.5638\\
		%ShapeSeg V2 &0.9046 &0.7930 &\textbf{0.7680} &\textbf{0.7671} &0.9275 &0.9273 \\
			SWN &\textbf{0.9101} &\textbf{0.8212} &\textbf{0.7516} &\textbf{0.7156} &0.7096 &0.7094 &\textbf{0.7365} &0.4256 &\textbf{0.9374} &\textbf{0.9370} &\textbf{0.8093} &\textbf{0.5645}\\
                \midrule
                %Improv (\%) &1.34	&3.47	&6.16	&10.22	&14.88	&16.86	&17.12	&22.19	&0.05	&0.07	&8.52	&10.00 \\ %absolute
                Improv (\%) &1.49	&4.41	&8.93	&16.66	&19.70	&20.58	&45.99	&83.53	&4.73	&4.74	&11.77	&21.53 \\
			\bottomrule
		\end{tabular}
	}
\end{table}

\vspace{-0.5cm}
%more balanced and robust overall performance 6.11\%  12.37\%
(1) We can see that our SWN variants significantly outperform four categories of TSC methods on three datasets with an average improvement of 9.28\% on BC tasks and 31.40\% on MC tasks. 
These results strongly testify the superiority of our method in dealing with the PSC problem especially on the condition of severe label sparsity issues.

(2) Our proposed ShapeWordNet surpasses the classical Shapelet-based ST and LS methods by a large margin, which underlines the significance of utilizing deep learning models to capture multivariate interaction and distilling both the local and global information for time series classification. 

(3) Compared with SAX-VSM and WEASEL+MUSE, our method wins overwhelmingly on all tasks. We believe this phenomenon concretely demonstrates the advantages of ShapeWords in representing discriminative local patterns over the SAX words and SFA words. Besides, it also indicates learning abstract feature representations is more effective than relying on discrete statistical patterns.

%more resistant to label sparsity compared with DL methods 14.14\% 
(4) In contrast to the CNN-based methods that rank the second in 5 tasks (3 for LSTM-FCN and 2 for TCN), our method works particularly better on three MC tasks by obtaining an average of 19.09\% rise in MAF1. Such a noticeable improvement soundly validates the effectiveness of leveraging ShapeWord Discretization and Cross-scale Contrastive Learning in label sparsity mitigation and invariant feature extraction.

\subsection{Ablation Study}
To evaluate the effectiveness of each component, we investigate the performance of three variants. Firstly, we observe that SWN w/o CCLM substantially surpasses the SWN w/o SD by approximately 17.92\% on CPSC MC, 12.16\% on Sleep MC and 109.77\%
on Georgia MC in MAF1, which substantiates the effectiveness of ShapeWord Discretization in reducing noise disturbance and distilling prominent features. In addition, the fact that SWN and SWN w/o SFI defeat SWN w/o CCLM on 10 tasks with at least 5.99\% rise of MC MAF1 tellingly verifies the remarkable progress made by the CCLM component in relieving the label sparsity. Moreover, the slight edges of SWN over SWN w/o SFI on 9 tasks indicate that SFI may contribute to promoting performance more or less.
%absolute up 9\%   6\%   19\% 4\% 

\subsection{ShapeWord Discretization Analysis}
As the results in Table \ref{table_2} imply, ShapeWord Discretization plays the most critical role in boosting the model's generalization. Therefore, it is necessary to explain how it really takes effect. To investigate the interpretability and effectiveness of ShapeWord Discretization, we take a close look at its characteristics through two visualization experiments. 

\begin{figure}[!htbp]
	\centering
	\vspace{-0.8cm}
	\setlength{\belowdisplayskip}{5cm}
	\subfigure[AF sample 1]
	{\label{m_1}
		\begin{minipage}[t]{0.5\linewidth}%0.25\textwidth
			%\centering
			\includegraphics[width=\linewidth]{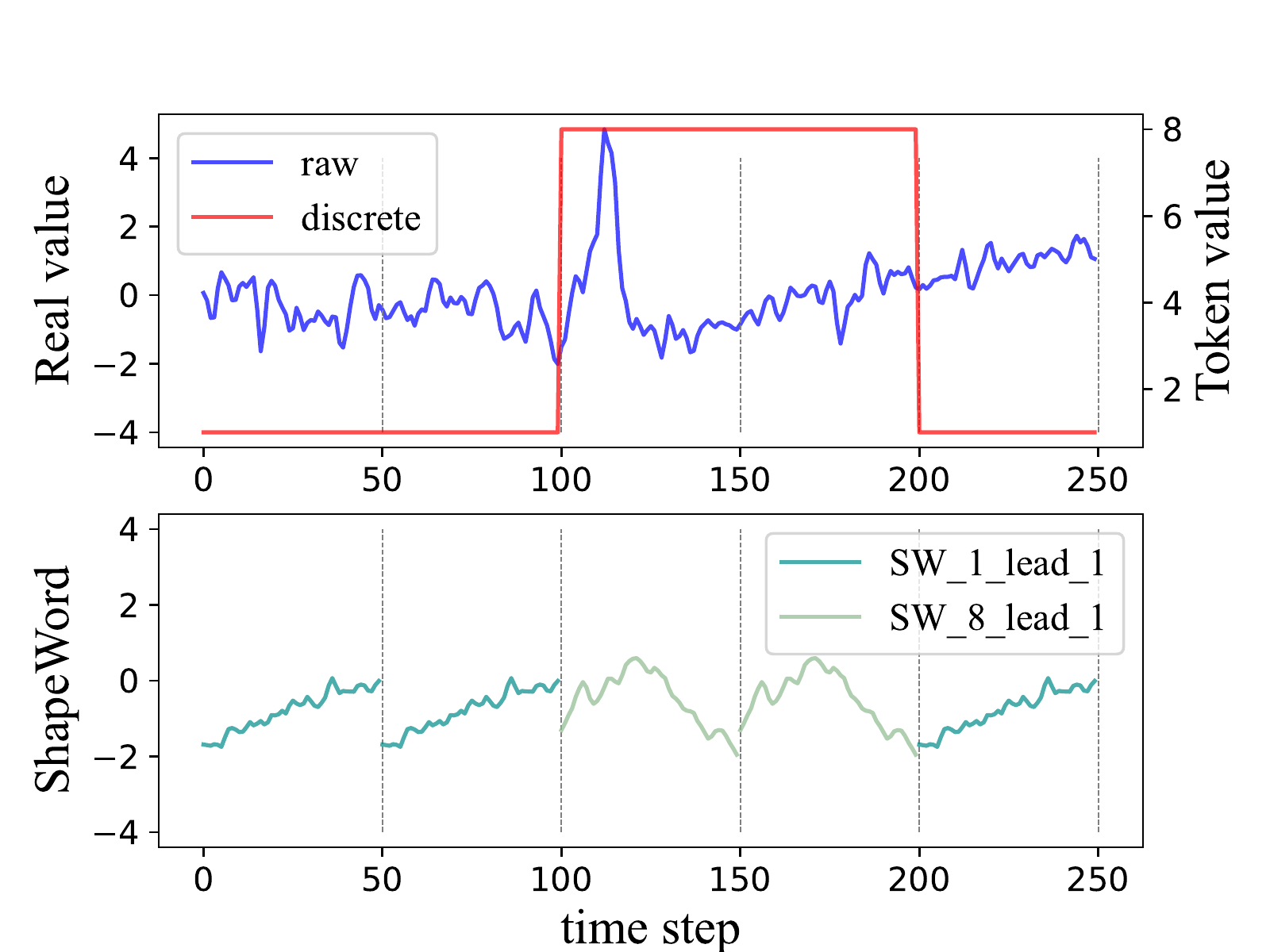}
   %{AF_9_shapelen_50_No_17_lead_1_train}
			%\caption{AF sample 1}
		\end{minipage}%
	}\subfigure[AF sample 2]{\label{m_2}
		\begin{minipage}[t]{0.5\linewidth}%0.25\textwidth
			%\centering
			\includegraphics[width=\linewidth]{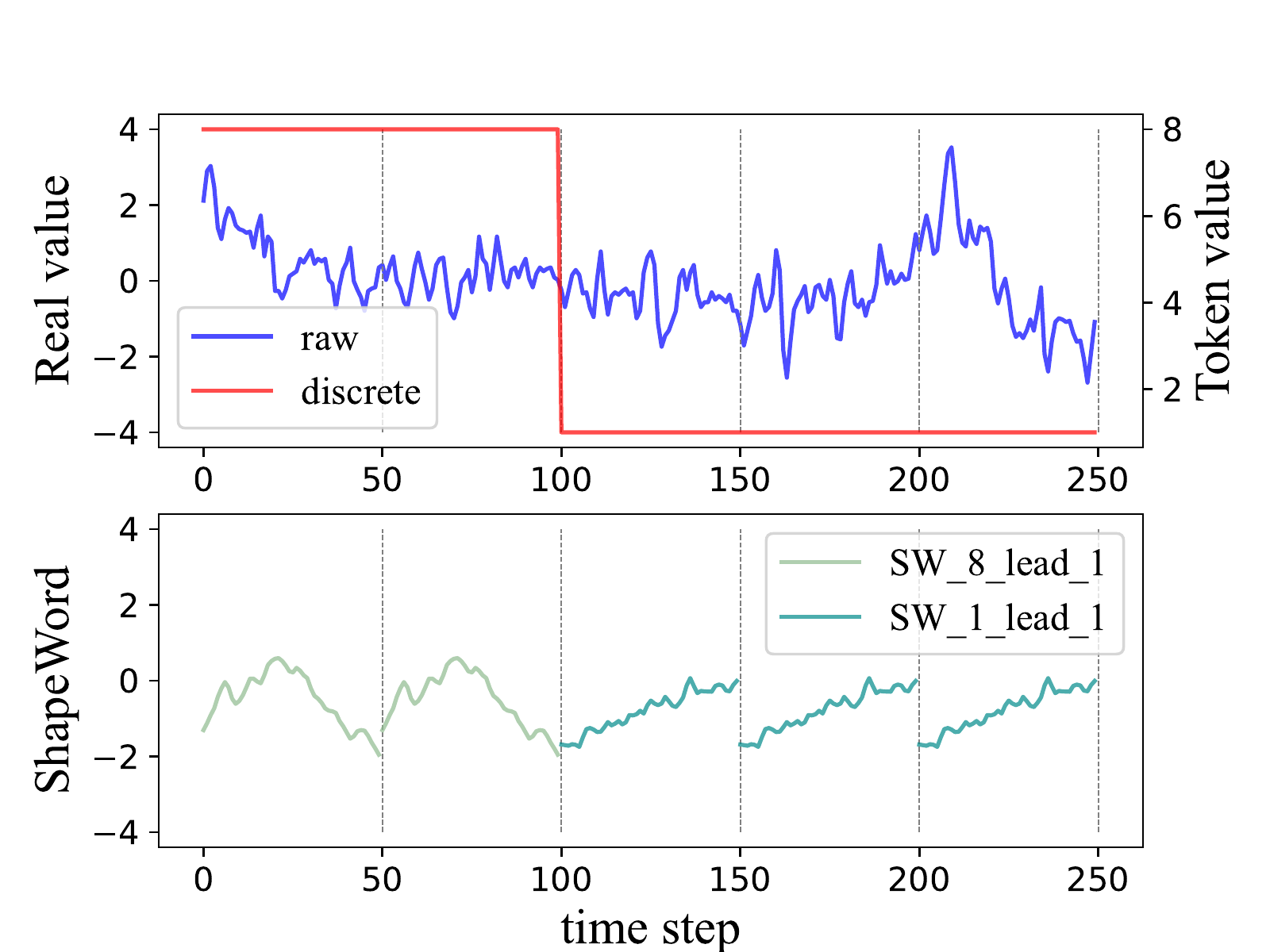}
   %{AF_9_shapelen_50_No_3_lead_1_train}
			%\caption{AF sample 2}
			%\caption{fig2}
		\end{minipage}%
	}%%\hfil 
        \vspace{-0.4cm}
	\subfigure[NSR sample 1]{\label{m_3}
		\begin{minipage}[t]{0.5\linewidth}%0.25\textwidth
			%\centering
			\includegraphics[width=\linewidth]{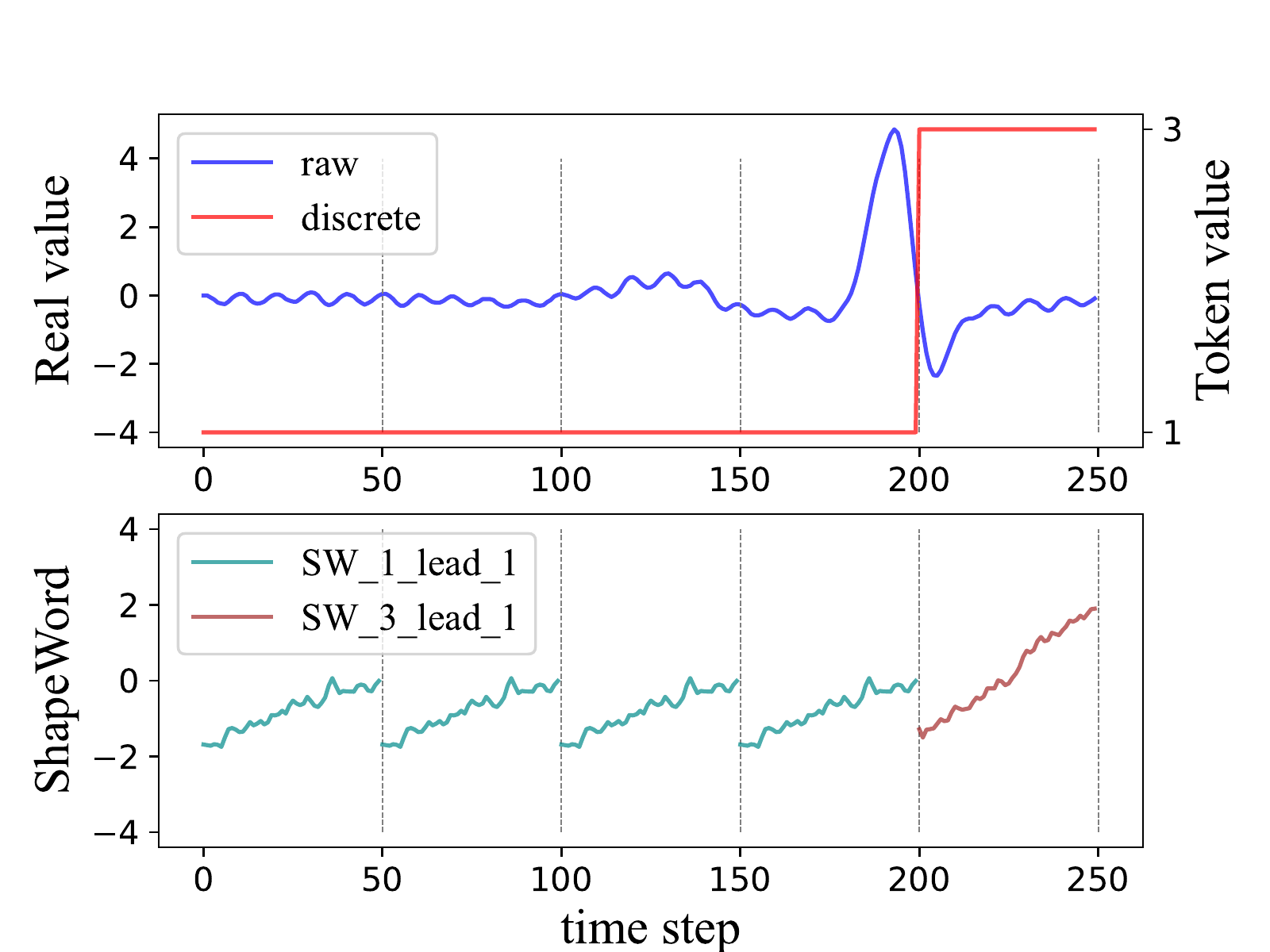}
   %{graphs/NSR_9_sample_1.pdf}
   %{NSR_9_shapelen_50_No_8_lead_1_train}
			%\caption{NSR sample 1}
		\end{minipage}%
	}\subfigure[NSR sample 2]{\label{m_4}
    	\begin{minipage}[t]{0.5\linewidth}%0.25\textwidth
    		%\centering
    		\includegraphics[width=\linewidth]{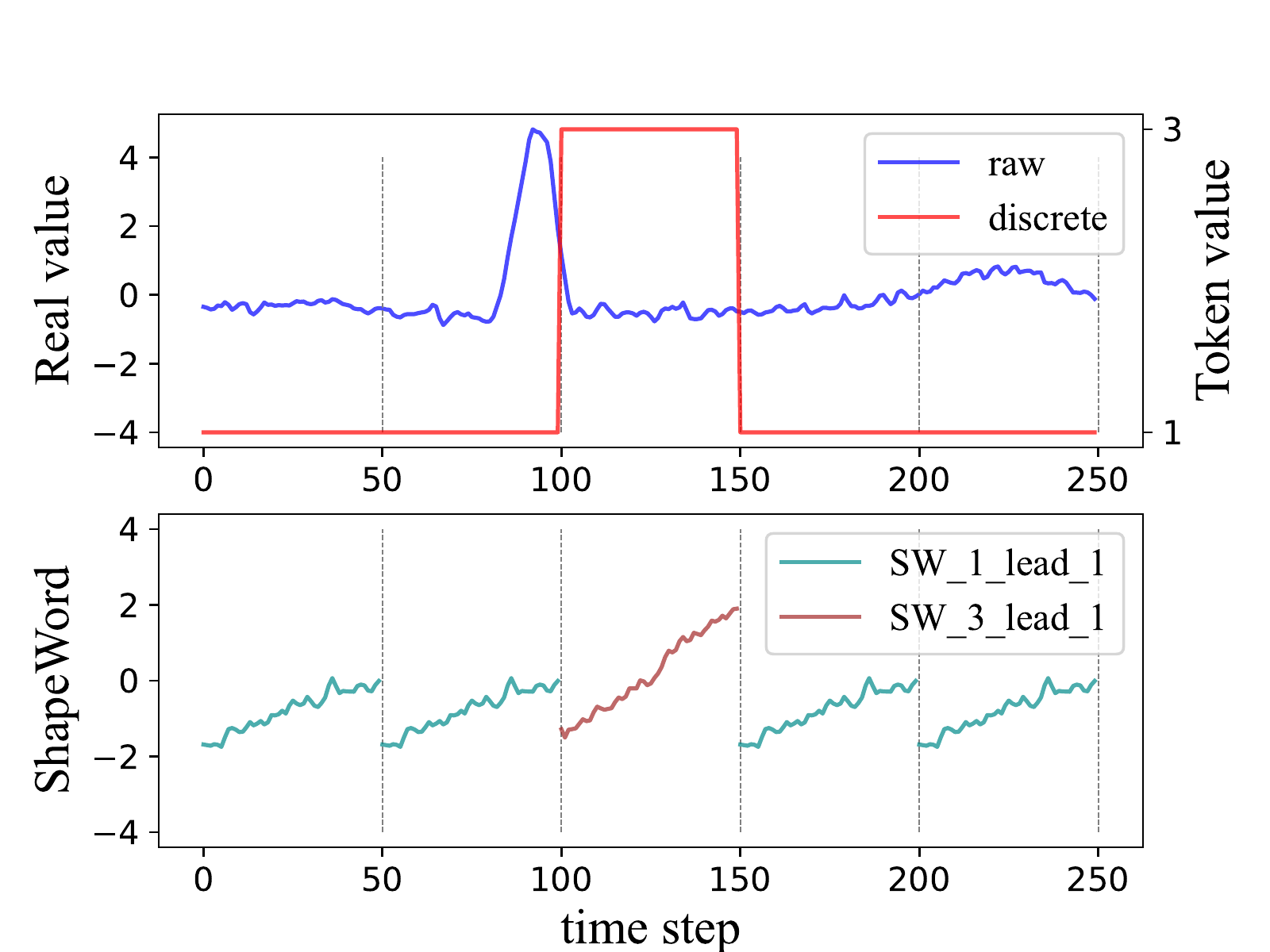}
      %{graphs/NSR_9_sample_2.pdf}
      %{NSR_9_shapelen_50_No_9_lead_1_train}
    		%\caption{NSR sample 2}
    	\end{minipage}%
    }%
   % \hfil
    \caption{A case of two AF samples and two NSR samples transformed by ShapeWord Discretization with $window\_size = 50$. The raw signals are in blue lines and their ShapeSentences are in red ones, where each token refers to a ShapeWord shown below.}
	\label{Fig_3}
\end{figure}

First, we conduct a case study to intuitively explore the interpretability of the ShapeWord Discretization. Figure \ref{Fig_3} illustrates a case of two atrial fibrillation (AF) samples and two sinus rhythm (NSR) samples, where the four raw signals in blue lines are discretized into four ShapeSentences ($[1,1,8,8,1]$, $[8,8,1,1,1]$, $[1,1,1,3,1]$ and $[1,1,3,1,1]$) in red lines respectively. It can be noticed that the pattern of AF, i.e. the disease of sustained tachyarrhythmia commonly seen in clinical practice, seems to be captured by the violently fluctuated ShapeWord $SW_8$ presented below. In contrast, the peculiarities of NSR, i.e. the normal state, seem to be represented by the combination of the ShapeWord $SW_1$ and $SW_3$. Hence, it is reasonable to believe that both the discriminative local patterns and their coherence can be well preserved via ShapeWord Discretization.

%\vspace{-0.7cm}

\begin{figure}[!htbp]
        \vspace{-0.5cm}
	\centering
	\setlength{\belowdisplayskip}{0.3cm}
	\subfigure[SWN w/o SD ]
	{\label{n_1}
		\begin{minipage}[htbp]{0.5\textwidth}%0.25\textwidth
			\centering
			\includegraphics[width=\linewidth]{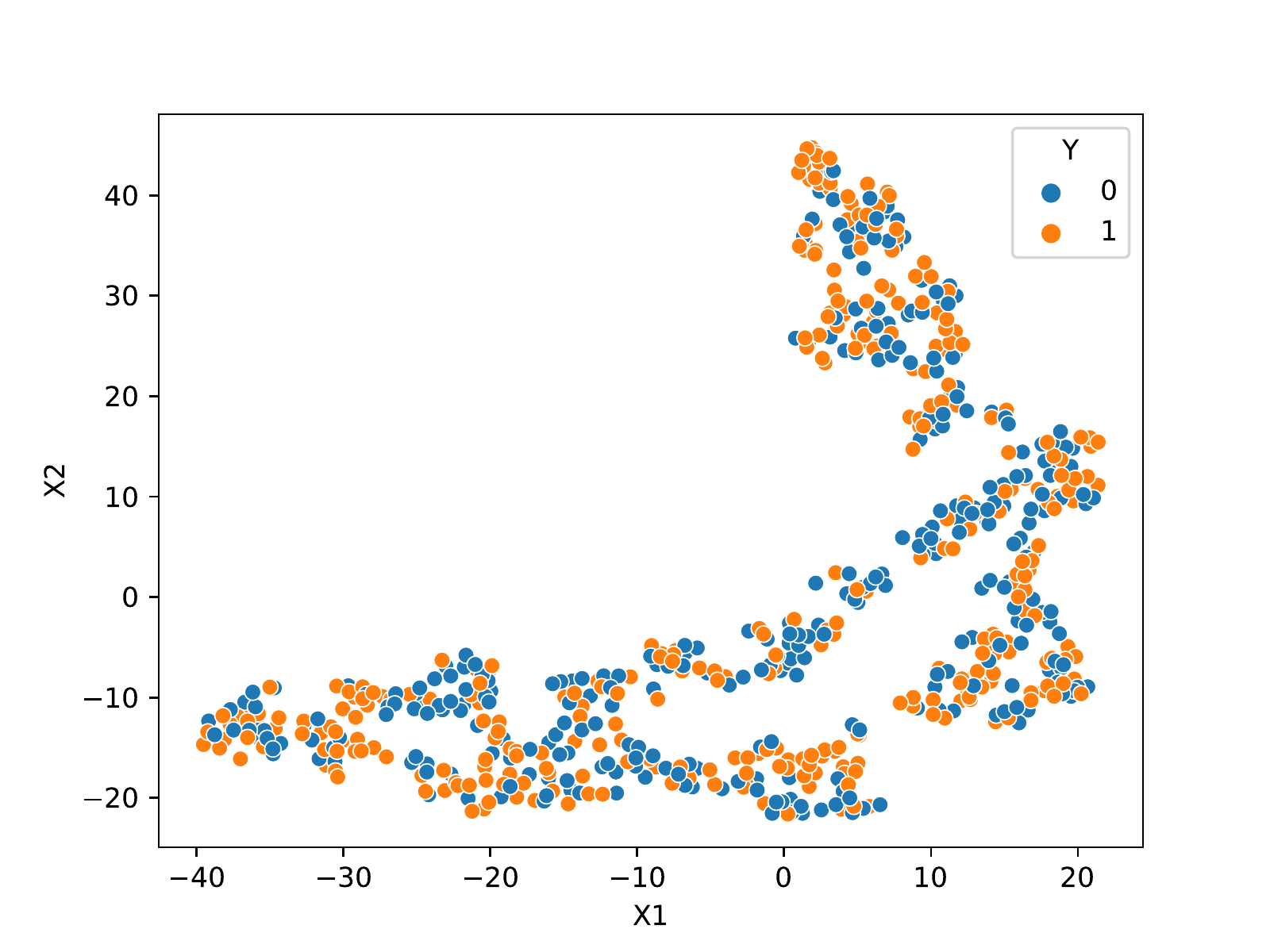}
			%\caption{AF sample 1}
		\end{minipage}
	}\subfigure[SWN w/o CCLM]{\label{n_2}
		\begin{minipage}[htbp]{0.5\textwidth}%0.25\textwidth
			\centering
			\includegraphics[width=\linewidth]{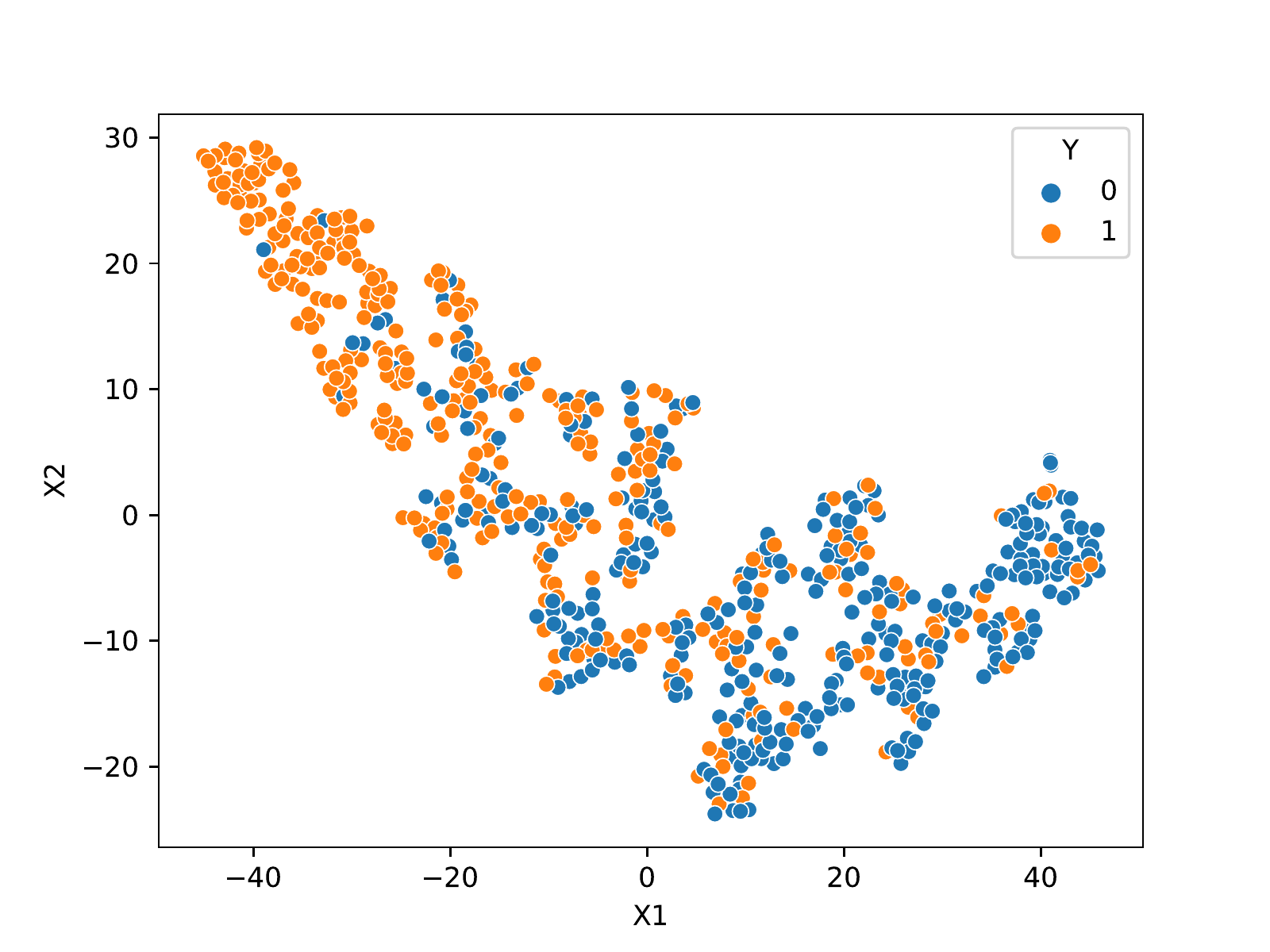}
			%\caption{AF sample 2}
			%\caption{fig2}
		\end{minipage}
	}
       \vspace{-0.3cm}
       \caption{T-SNE visualization of representations produced by
(a) SWN w/o SD and (b) SWN w/o CCLM on Georgia BC task (0 indicates positive and 1 means negative).}
	\label{Fig_5}
\end{figure}

\vspace{-0.3cm}
Second, to vindicate the contributions of ShapeWord Discretization to DL model's representation learning, we conduct the t-SNE analysis \cite{van2008visualizing} to compare the representations output by SWN w/o CCLM and SWN w/o SD respectively on the Georgia BC task.
In Figure \ref{n_1} and \ref{n_2}, the sample representations output by SWN w/o CCLM (i.e. the variant with the ShapeWord Discretization) are more closely grouped with a clearer clustering boundary than those output by the SWN w/o SD, which defensibly testifies the effect of ShapeWord Discretization on relieving the influence of data noise and promoting the representational ability of deep neural networks.

\vspace{-0.3cm}
\subsection{Hyper-parameter Sensitivity}
In this section, we discuss the impact of several key hyper-parameters on the performance of our method, which includ the Scale Number $N_s$, the Loss Balance Factor $\lambda$, the ShapeWord Number $N_{SW}$ and the ShapeWord Length $L_{SW}$.

\vspace{-0.7cm}
\subsubsection{Performance w.r.t. Scale Number.}
We display the model performance w.r.t. the scale number $N_s$ in  Figure \ref{Fig_6}, where the scales for MST are successively picked from the range [5,10,25,50,100], e.g. 2 for [5,10] and 3 for [5,10,25].  
Figure \ref{s_1} and Figure \ref{s_2} show that for different datasets the impact of scale number is different and an ideal interval of this parameter shared by three datasets seems to be around [2,3]. Based on this observation, we implement 3 scales of MST for SWN in our experiments.

\vspace{-0.3cm}
\subsubsection{Performance w.r.t. Loss Balance Factor.}
\begin{figure}[!htbp]
       \vspace{-0.5cm}
       \setlength{\belowdisplayskip}{3cm}
	\centering
	\subfigure[BC w.r.t. $N_s$]
	{        \label{s_1}
			\centering
			\includegraphics[width=0.24\textwidth]{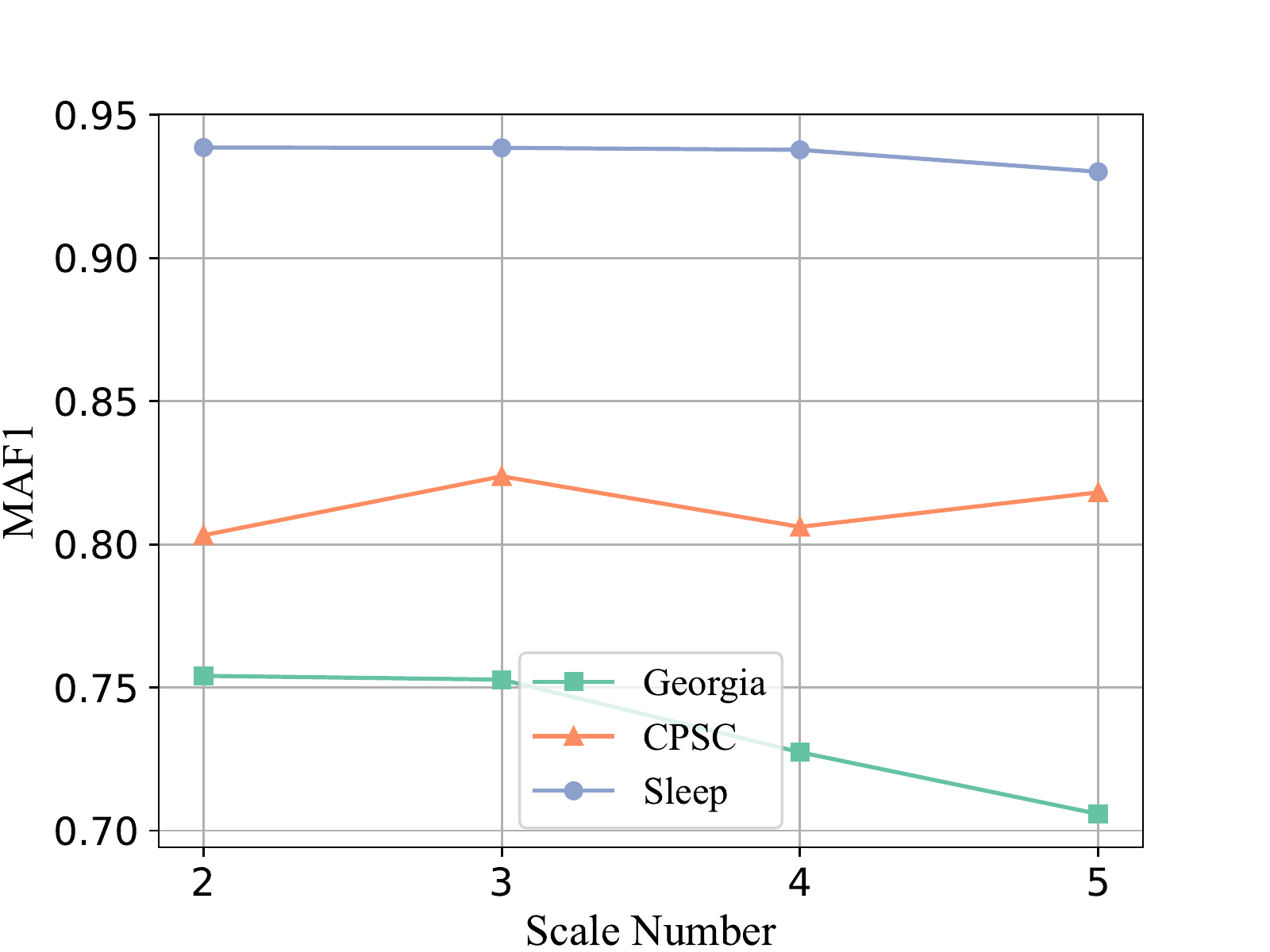}
			%\caption{AF sample 1}
	}%
	%\quad
	\subfigure[MC w.r.t. $N_s$]{\label{s_2}
			\centering
			\includegraphics[width=0.24\textwidth]{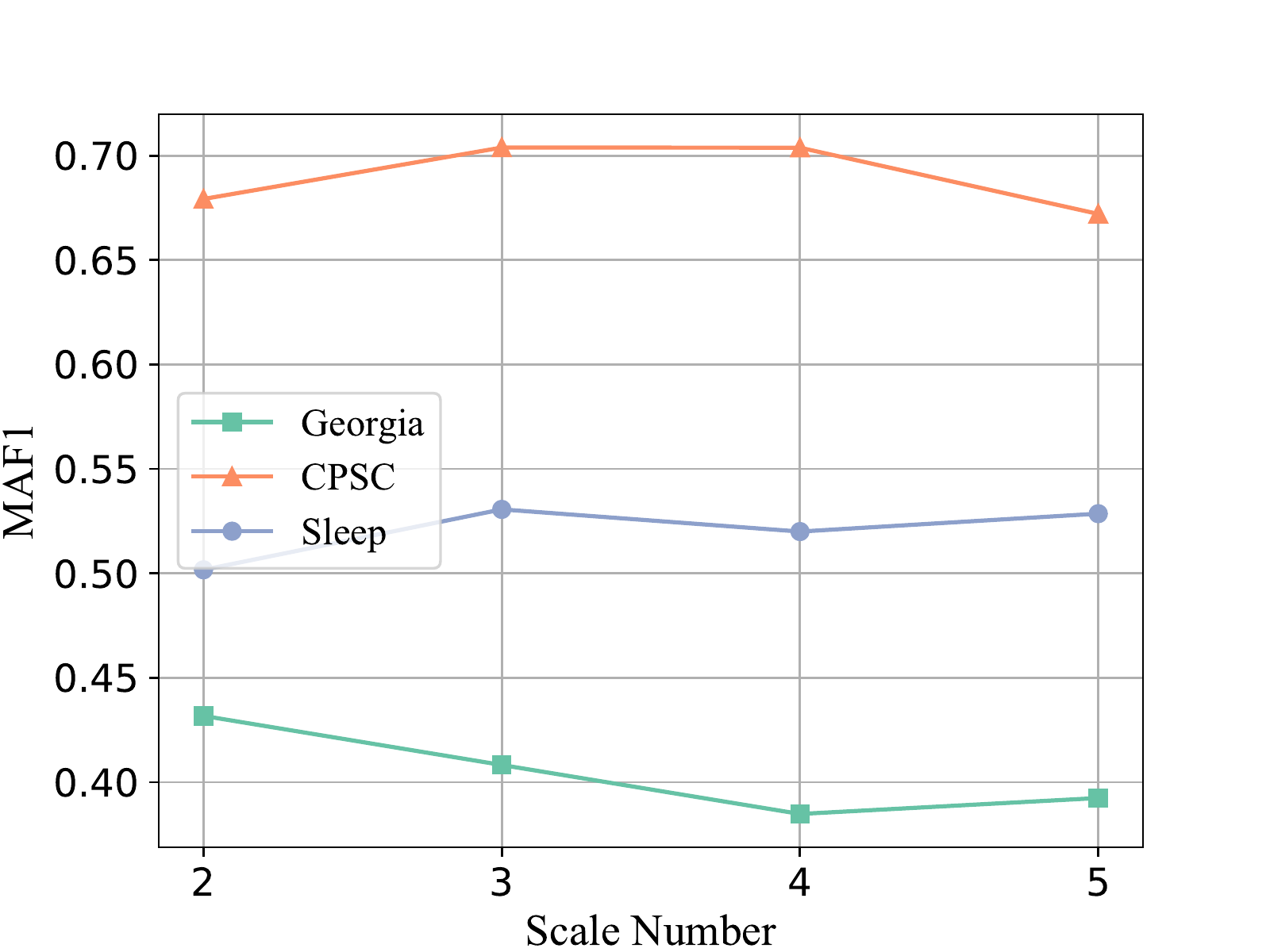}
			%\caption{AF sample 2}
			%\caption{fig2}
	}%
        \subfigure[BC w.r.t. $\lambda$]
	{\label{q_1}
		%0.25\textwidth
			\centering
			\includegraphics[width=0.24\textwidth]{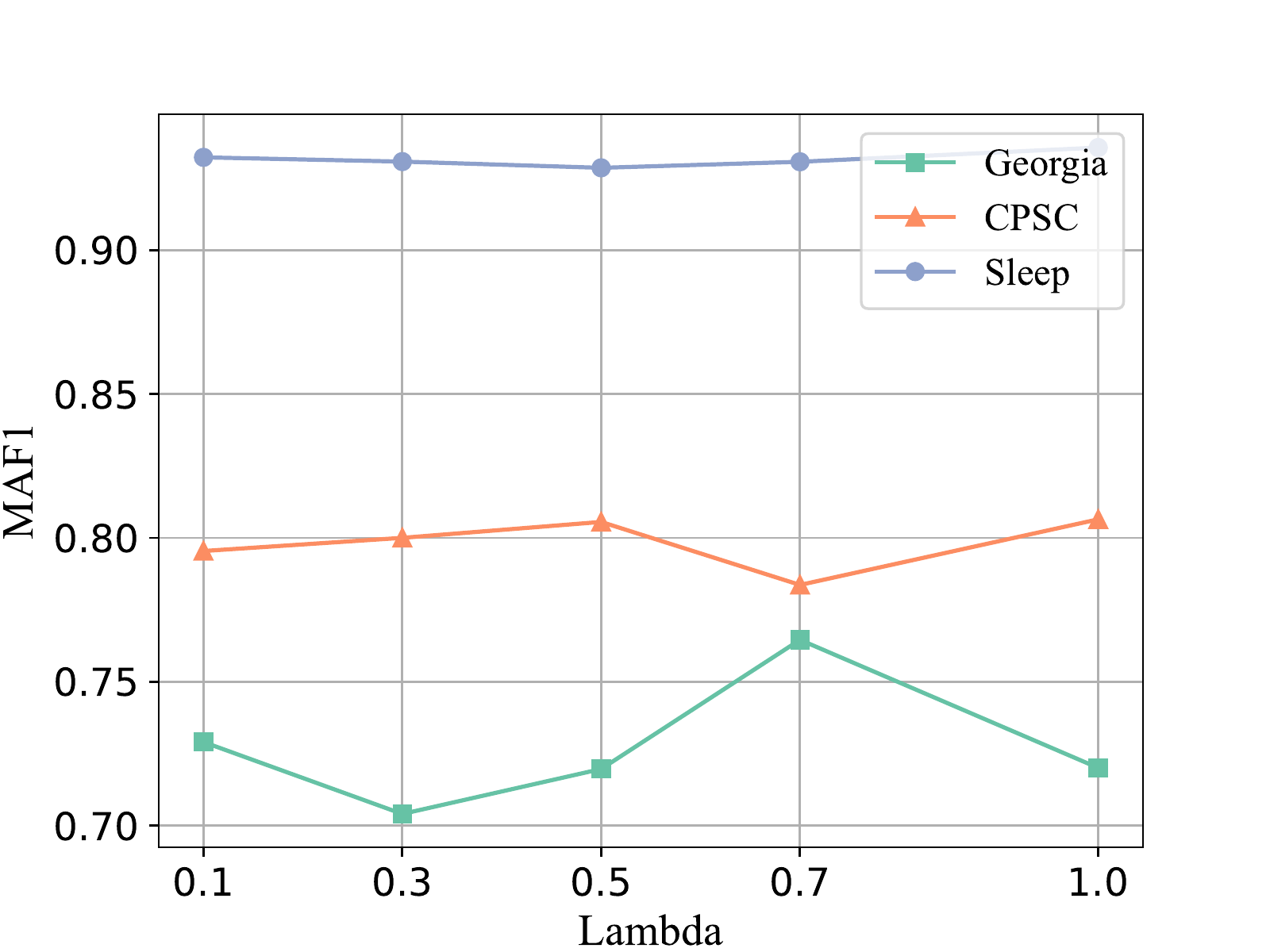}
			%\caption{AF sample 1}
	}%
	%\quad
	\subfigure[MC w.r.t. $\lambda$]{
                \label{q_2}
			\centering
			\includegraphics[width=0.24\textwidth]{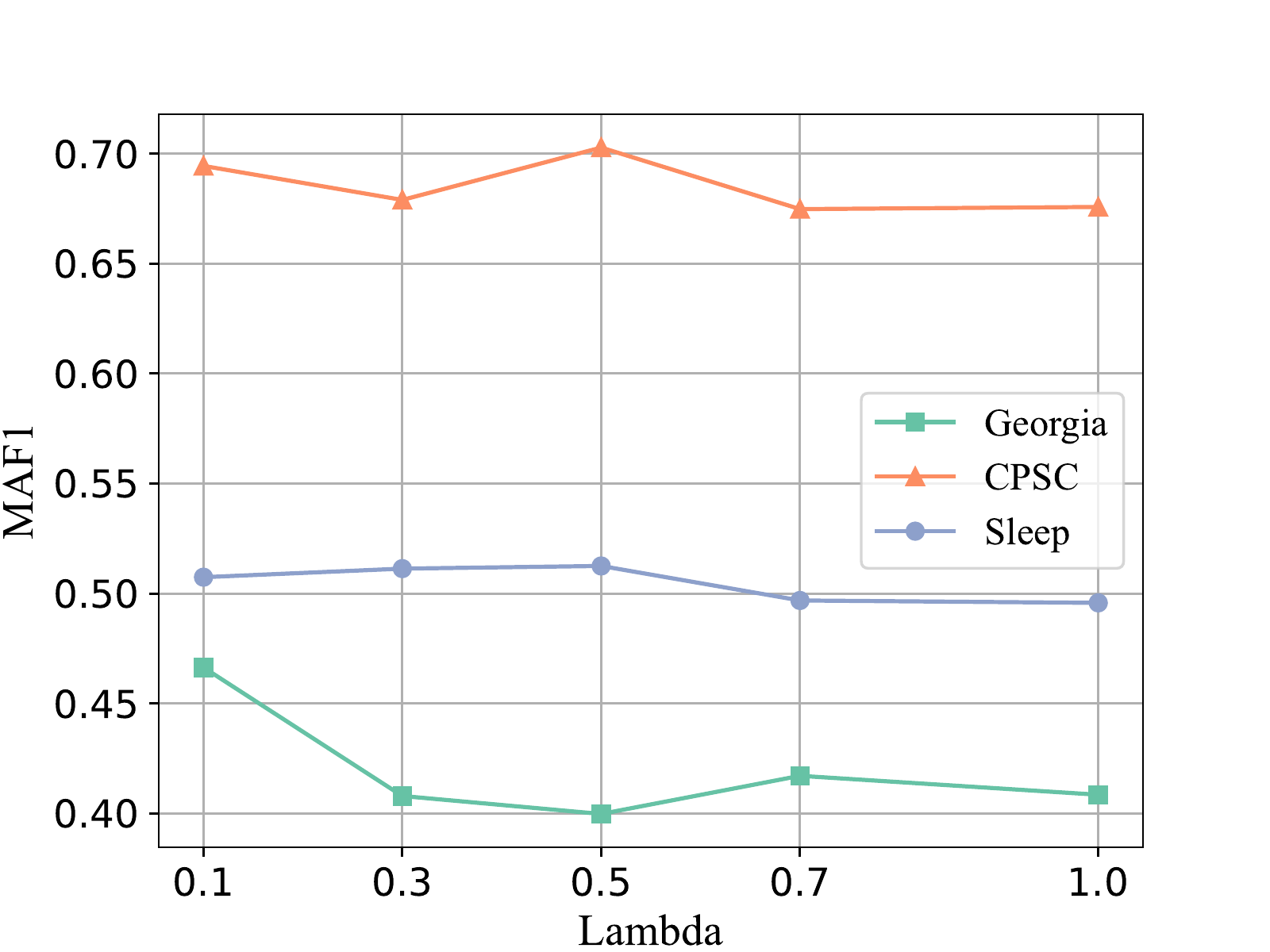}
			%\caption{AF sample 2}
			%\caption{fig2}
       }
       \vspace{-0.3cm}
       \caption{SWN's performance of MAF1 w.r.t. the number of scales $N_s$ and the loss balance factor $\lambda$ on three datasets over BC task and MC task.}
	\label{Fig_6}
\end{figure}

%\subsubsection{$\lambda$}
The parameter $\lambda$ in Equation \ref{eq_10} is to balance the loss between classification training and contrastive learning. As is illustrated in Figure  \ref{q_1} and \ref{q_2}, the performance of SWN seems more likely to be affected by $\lambda$ on MC tasks than on BC tasks, and the overlapping optimal interval of $\lambda$ for three datasets is suggested to be $[0.3,0.7]$. In our experiment, we adopt $\lambda=0.5$ as default given the fact that SWN obtains the best performance on two datasets under this condition. 

\vspace{-1cm}
\subsubsection{Performance w.r.t. ShapeWord Number.}
\begin{figure}[!htbp]
        \vspace{-0.8cm}
	\centering
        \subfigure[$L_{SW}= 5$]
	{\label{p_0}
		\begin{minipage}[htbp]{0.32\linewidth}%0.25\textwidth
			%\centering
			\includegraphics[width=\linewidth]{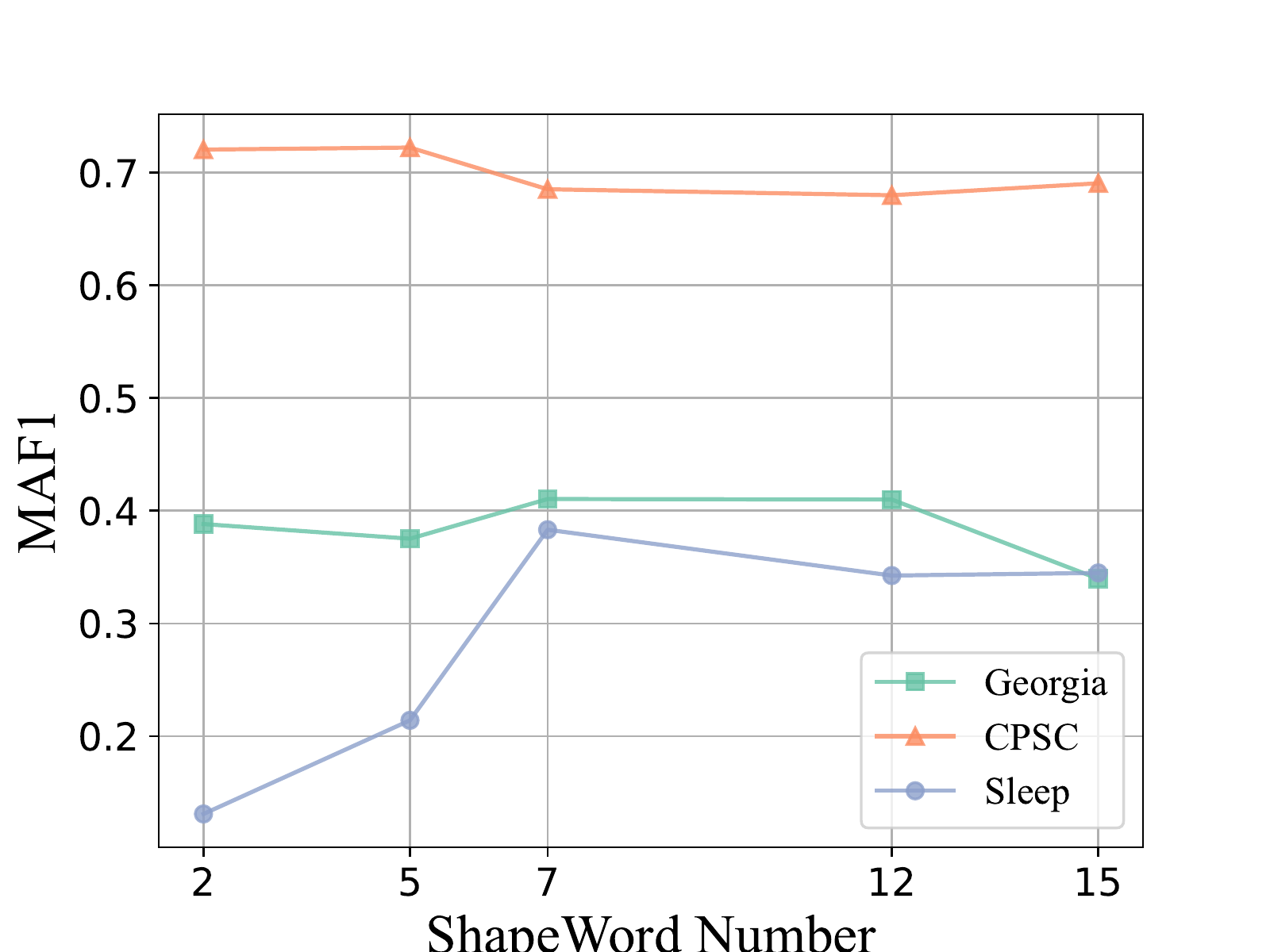}
   %{AF_9_shapelen_50_No_17_lead_1_train}
			%\caption{AF sample 1}
		\end{minipage}%
	}\subfigure[$L_{SW}= 10$]
	{\label{p_1}
		\begin{minipage}[htbp]{0.32\linewidth}%0.25\textwidth
			%\centering
			\includegraphics[width=\linewidth]{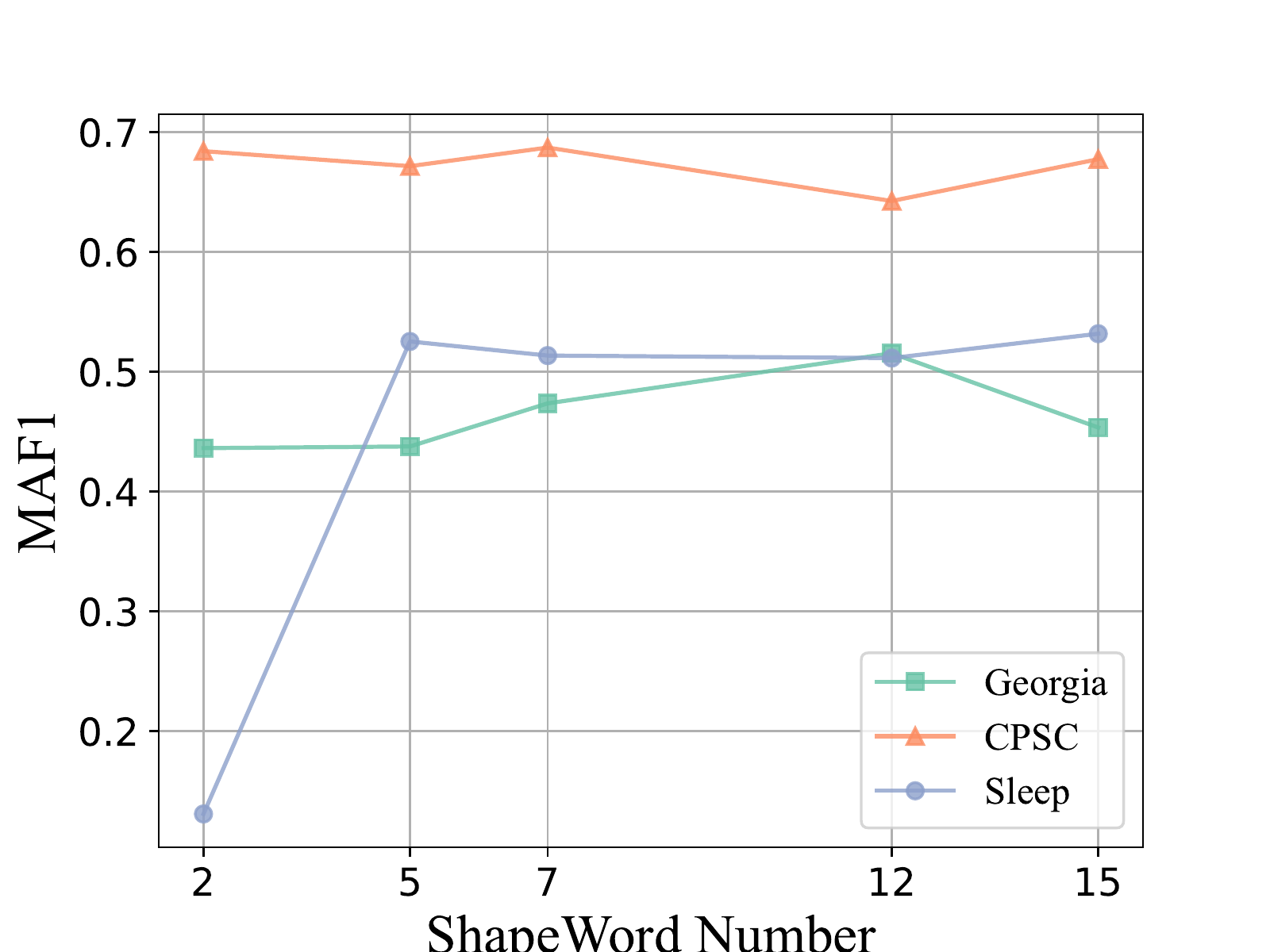}
   %{AF_9_shapelen_50_No_17_lead_1_train}
			%\caption{AF sample 1}
		\end{minipage}%
	}
       \subfigure[$L_{SW}= 25$]{\label{p_2}
		\begin{minipage}[htbp]{0.32\linewidth}%0.25\textwidth
			%\centering
			\includegraphics[width=\linewidth]{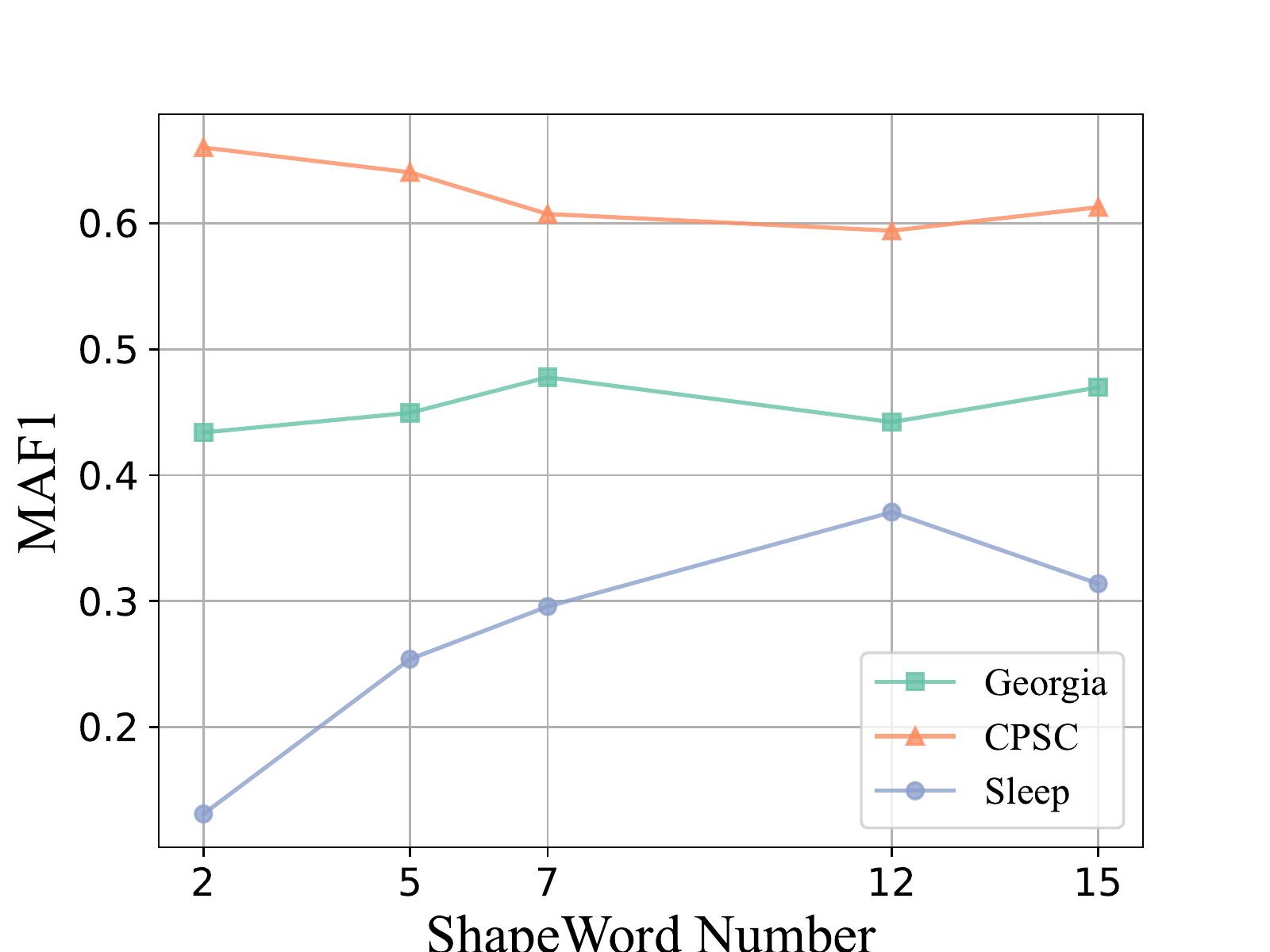}
   %{AF_9_shapelen_50_No_3_lead_1_train}
			%\caption{AF sample 2}
			%\caption{fig2}
		\end{minipage}%
	}%%\hfil  %\vspace{-0.4cm}
       \vspace{-0.4cm}
 \subfigure[$L_{SW}= 50$]{\label{p_3}
		\begin{minipage}[htbp]{0.32\linewidth}%0.25\textwidth
			%\centering
			\includegraphics[width=\linewidth]{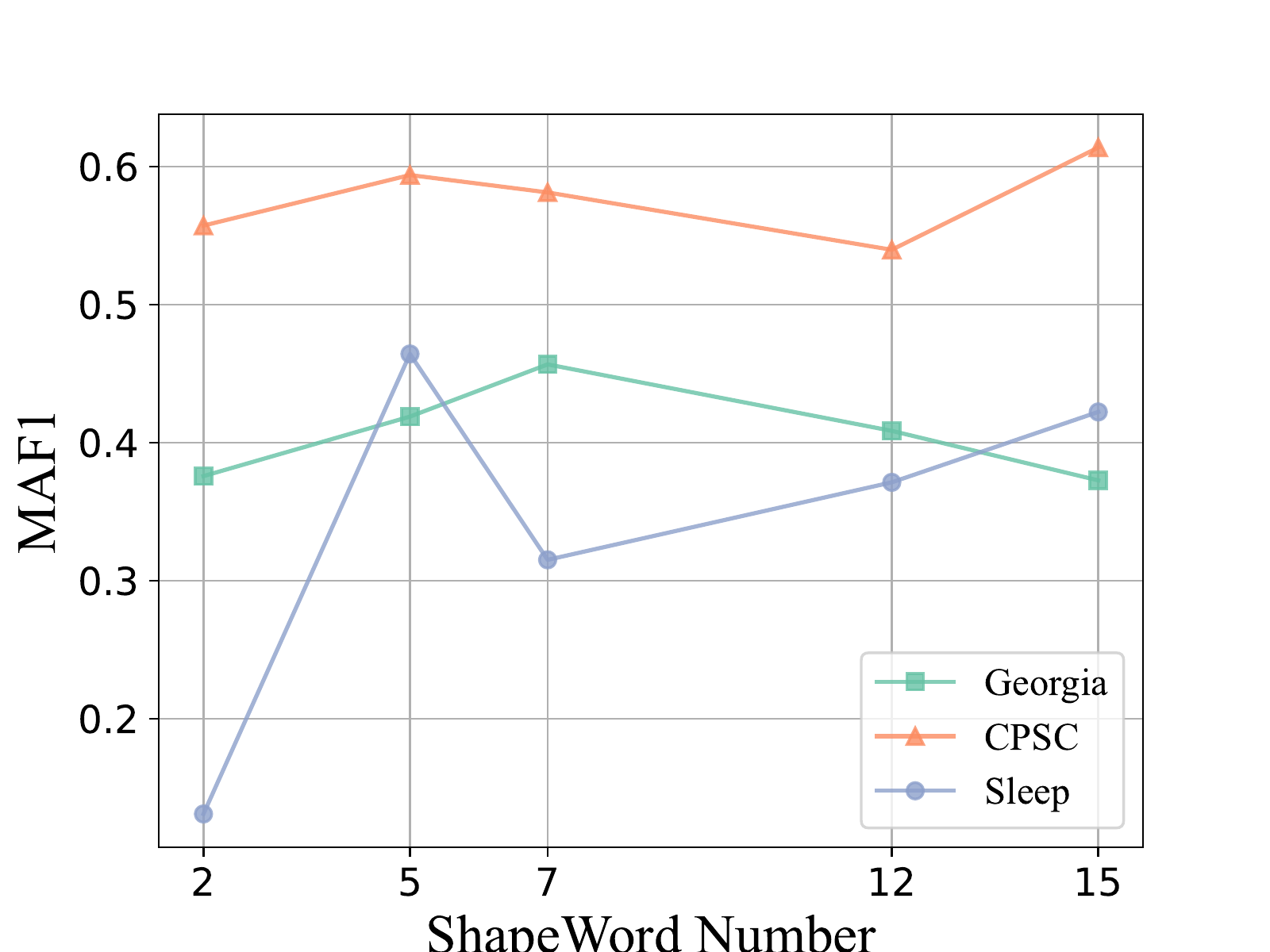}
   %{NSR_9_shapelen_50_No_8_lead_1_train}
			%\caption{NSR sample 1}
		\end{minipage}%
	}
        \subfigure[$L_{SW}= 100$]{\label{p_4}
    	\begin{minipage}[htbp]{0.32\linewidth}%0.25\textwidth
    		%\centering
    		\includegraphics[width=\linewidth]{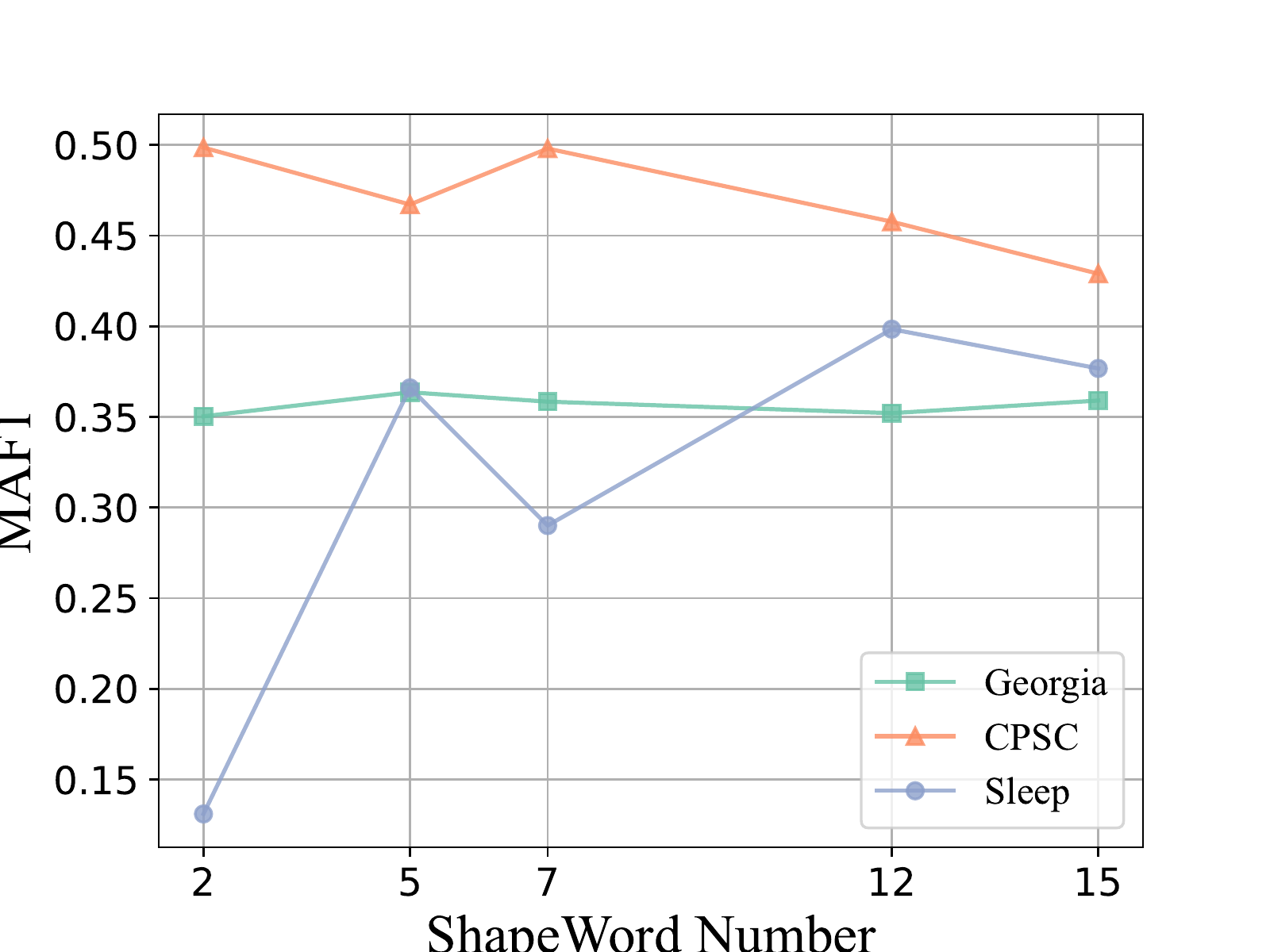}
      %{NSR_9_shapelen_50_No_9_lead_1_train}
    		%\caption{NSR sample 2}
    	\end{minipage}%
    }%
   % \hfil
    \vspace{-0.3cm}
    \caption{Performance of SWN w/o CCLM w.r.t. ShapeWord Number in MC tasks under five different settings of ShapeWord Length.}
	\label{Fig_8}
\end{figure}

Since  ShapeWord is defined as the centroid of a cluster of similar shapelets, its number hence depends on the cluster number. In this article, we leverage K-means to generate ShapeWords. To exclude the impact of contrastive learning, we adopt the variant SWN w/o CCLM to conduct the sensitivity experiment regarding the parameter of ShapeWord Number $N_{SW}$.
Graphs in Figure \ref{Fig_8} show that despite the influence of ShapeWord Length $L_{SW}$, the overlapping optimal interval of three datasets for $N_{SW}$ is approximately $[5,12]$. In our experiment, we choose the class number as this parameter's default setting for each dataset (i.e. 8 for Sleep, 9 for CPSC and Georgia), which is also consistent with the original definition of shapelets.

\vspace{-1cm}
\subsubsection{Performance w.r.t. ShapeWord Length.}
It can be observed from Figure \ref{Fig_8} that the performance curve of each dataset shows a trend of first moving up and then going down as $L_{SW}$ increases, which indicates short ShapeWords are more suitable for physiological signal discretization. And the overlapping optimal interval of ShapeWord Length among three datasets is about $[10,50]$, which is why we choose the three scales $[10,25,50]$ for SWN.

\vspace{-0.3cm}
\section{Conclusion}
\vspace{-0.1cm}
In this paper, we proposed ShapeWordNet to deal with physiological signal classification. The uniqueness of our model is to generate prototypes of discriminative local patterns via shapelets and to discretize point-wise raw signals into token sequences of subseries. Given the label sparsity issue, we designed a cross-scale contrastive learning mechanism to assist model optimization, where a multi-scale ShapeSentence transformation strategy was adaptively utilized to augment the data. The experimental results demonstrated both the effectiveness and the interpretability of our method, paving the way for its extension to general time series analysis in the furture.

\vspace{-0.5cm}
\subsubsection{Acknowledgement.} This research was partially supported by grant from the National Natural Science Foundation of China (Grant No. 61922073). This work also thanks the support of fundings MAI2022C007 and WK5290000003.

%\clearpage
\bibliographystyle{splncs04}
\bibliography{ref}

\end{document}